\pdfoutput=1
\documentclass[twoside]{article}

\usepackage[accepted]{aistats2021}
\usepackage{sidecap}
\usepackage{hyperref}
\usepackage{graphicx}
\usepackage{booktabs}
\usepackage{subcaption}
\usepackage{float}
\usepackage{color}
\usepackage{algorithm}
\usepackage{algorithmic}
\usepackage{amsmath}
\usepackage{bbm}
\usepackage{enumitem}
\usepackage{bm,amsbsy} 
\usepackage{amsfonts,amsmath,amssymb,amsthm,mathtools}
\usepackage{float}
\begin{document}

\newcommand{\inv}{^{-1}}
\newcommand{\trp}{{^\top}} 
\newcommand{\onehlf}{\frac{1}{2}}
\newcommand{\tonehlf}{\tfrac{1}{2}}
\renewcommand{\eqref}[1]{eq.~\ref{eq:#1}}
\newcommand{\Nrm}{\mathcal{N}}
\newcommand{\Tr}{\mathrm{Tr}}
\newcommand{\diag}{\mathrm{diag}}
\newcommand{\figref}[1]{Fig.~\ref{fig:#1}}  
\newcommand{\secref}[1]{Sec.~\ref{sec:#1}}  
\newcommand{\tabref}[1]{Table ~\ref{tab:#1}}  
\newcommand{\algoref}[1]{Algorithm~\ref{algo:#1}}  

\newcommand{\propref}[1]{Prop.~\ref{prop:#1}}

\newcommand{\suppsecref}[1]{Appendix Sec.~\ref{supp:#1}}  

\newcommand{\vx}{\mathbf{x}}
\newcommand{\var}{\mathrm{var}}
\newcommand{\Dat}{\mathcal{D}}

\newcommand{\Ev}{\mathcal{E}}
\newcommand{\vxloc}{\mathcal{X}}
\newcommand{\vkx}{\mathbf{k_x}}
\newcommand{\vv}{\mathbf{v}}
\newcommand{\vky}{\mathbf{k_y}}
\newcommand{\vf}{\mathbf{f}}
\newcommand{\vb}{\mathbf{b}}
\newcommand{\valpha}{\mathbf{\ensuremath{\bm{\alpha}}}}
\newcommand{\vbeta}{\mathbf{\ensuremath{\bm{\beta}}}}
\newcommand{\vphi}{\mathbf{\ensuremath{\bm{\phi}}}}
\newcommand{\bmu}{\mathbf{\ensuremath{\bm{\mu}}}}
\newcommand{\vu}{\mathbf{u}}
\newcommand{\vk}{\mathbf{k}}
\newcommand{\bK}{\mathbf{K}}
\newcommand{\Vm}{\mathbb{V}} 
\newcommand{\vone}{\mathbf{1}} 
\newcommand{\vm}{\mathbf{m}}
\newcommand{\vz}{\mathbf{z}}
\newcommand{\vw}{\mathbf{w}}
\newcommand{\vr}{\mathbf{r}} 
\newcommand{\vsbold}{\mathbf{s}}
\newcommand{\vs}{\mathbf{s}} 
\newcommand{\fmap}{\bm{\lambda}{_{map}}}
\newcommand{\bphi}{\bm{\phi}}
\newcommand{\phimap}{{\hat {\bm{\phi}}}{_{map}}}
\newcommand{\fmapstar}{\mathbf{f^*_{map}}}
\newcommand{\muf}{\ensuremath{\mu_f}}
\newcommand{\vmuf}{\mathbf{\ensuremath{\bm{\mu}_f}}}
\newcommand{\vpi}{\mathbf{\ensuremath{\bm{\pi}}}}
\newcommand{\vmu}{\mathbf{\ensuremath{\bm{\mu}}}}
\newcommand{\vtheta}{\mathbf{\ensuremath{\bm{\theta}}}}
\newcommand{\veta}{\mathbf{\ensuremath{\bm{\eta}}}}
\newcommand{\LL}{\ensuremath{\mathcal{L}}}
\newcommand{\vlam}{\bm{\lambda}}
\newcommand{\mR}{\mathbf{R}}
\newcommand{\mW}{\mathbf{W}}
\newcommand{\mZ}{\mathbf{Z}}
\newcommand{\mC}{\mathbf{C}}

\newcommand{\vX}{\mathcal{X}}
\newcommand{\vK}{\mathbf{K}}
\newcommand{\mJ}{\mathbf{J}}

\newcommand{\set}[1]{\{#1\}}
\newcommand{\kml}{{\hat \vk}_{ML}}
\newcommand{\vy}{\mathbf{y}}
\newcommand{\vh}{\mathbf{h}}
\newcommand{\vc}{\mathbf{c}}
\newcommand{\vp}{\mathbf{p}}
\newcommand{\vg}{\mathbf{g}}
\newcommand{\Lprior}{\Lambda_p}
\newcommand{\Lix}{L_{x}}
\newcommand{\thetmap}{\theta_{\mu}}
\newcommand{\Lmap}{\Lambda_\mu}
\newcommand{\thetli}{\theta_{l}}
\newcommand{\Lli}{\Lambda_{l}}
\newcommand{\nsevar}{\sigma^2}
\newcommand{\nsestd}{\sigma}
\newcommand{\vchi}{\bm{\chi}}
\newcommand{\vomega}{\bm{\omega}}
\newcommand{\tr}{^\top}

\newtheorem{thm}{Theorem}[section]
\newtheorem{lem}{Lemma}[section]
\newtheorem{prop}{Proposition}
\newtheorem{defn}{Definition}[section]
\newtheorem{condition}{Condition}
\def\argmax{\mathop{\rm arg\,max}}
\def\argmin{\mathop{\rm arg\,min}}

\def\Pr{\ensuremath{\text{Pr}}}
\newcommand{\mj}[1]{{\color{blue}{ mijung : #1}}}

%

%

\twocolumn[

\aistatstitle{Dirichlet Pruning for Neural Network Compression}

\aistatsauthor{ Kamil Adamczewski \And Mijung Park}

\aistatsaddress{ Max Planck Institute for Intelligent Systems \\ ETH Zürich \And  Mac Planck Institute for Intelligent Systems \\ University of Tübingen } ]

\begin{abstract}
We introduce \textit{Dirichlet pruning}, a novel  post-processing technique to transform a large neural network model into a compressed one.
Dirichlet pruning is a form of structured pruning which assigns the Dirichlet distribution over each layer's  channels in convolutional layers (or neurons in fully-connected layers), and estimates the parameters of the distribution over these units using variational inference.
The learned distribution allows us to remove unimportant units, resulting in a compact architecture containing only crucial features for a task at hand. 
%
The number of newly introduced Dirichlet parameters is only linear in the number of channels, which allows for rapid training, requiring as little as one epoch to converge. 
We perform extensive experiments, in particular on larger architectures such as VGG and ResNet (45\% and 58\% compression rate, respectively) where our method achieves the state-of-the-art compression performance and provides interpretable features as a by-product. 
\end{abstract}

\section{INTRODUCTION}

Neural network models have 
achieved state-of-the art results in various tasks, including object recognition and reinforcement learning \cite{1211479, GU2018354, mnih2013playing, achanta2012slic, felzenszwalb2009object}. The algorithmic and hardware advances propelled the network sizes which have increased several orders of magnitude, from the LeNet \cite{lecun1998gradient} architecture with a few thousand parameters to ResNet \cite{he2016deep} architectures with almost 100 million parameters. Recent language models require striking 175 billion parameters \cite{brown2020language}.

However, large architectures incur high computational costs and memory requirements at both training and test time. They also become hard to analyze and interpret. 
%
Besides, it is unclear
whether a network needs all the parameters given by a hand-picked, rather than intelligently-designed architecture. 
For example, VGG-16 \cite{simonyan2014very} consists of layers containing 64, 128, 256, and 512 channels, respectively. However, there is no evidence that all those channels are necessary for maintaining the model's generalization ability.
%


Previous work noticed and addressed these redundancies in neural network architectures \cite{lecun1990optimal, hassibi1993second}. Subsequently, neural network compression became a  popular research topic, proposing smaller, slimmer, and faster networks while maintaining little or no loss in the immense networks' accuracy
\cite{hinton2015distilling, howard2017mobilenets, iandola2016squeezenet}. 
However, many of existing approaches 
judge the importance of weight parameters relying on
the proxies such as weights' magnitude in terms of L1 or L2 norms \cite{he2019filter}. 
In this work, we take a different route by learning the importance of a computational unit, a channel in convolutional layers or a neuron in fully connected layers. For simplicity, we will use the term, channels, as removable units throughout the paper, with a focus on convolutional neural networks (CNNs).
%
%


Our pruning technique provides a numerical way to compress the network by introducing a new and simple operation per layer to existing neural network architectures. These operations capture the relative importance of each channel to a given task.
%
%
%
We remove the channels with low importance to obtain a compact representation of a network as a form of structured pruning.

The learned importance of channels also naturally provides a ranking among the channels in terms of their significance. Visualizing the feature maps associated with highly-ranked channels provides intuition why compression works and what information is encoded in the remaining channels after pruning.

\begin{figure*}[t]
\begin{center}
\includegraphics[scale=0.55]{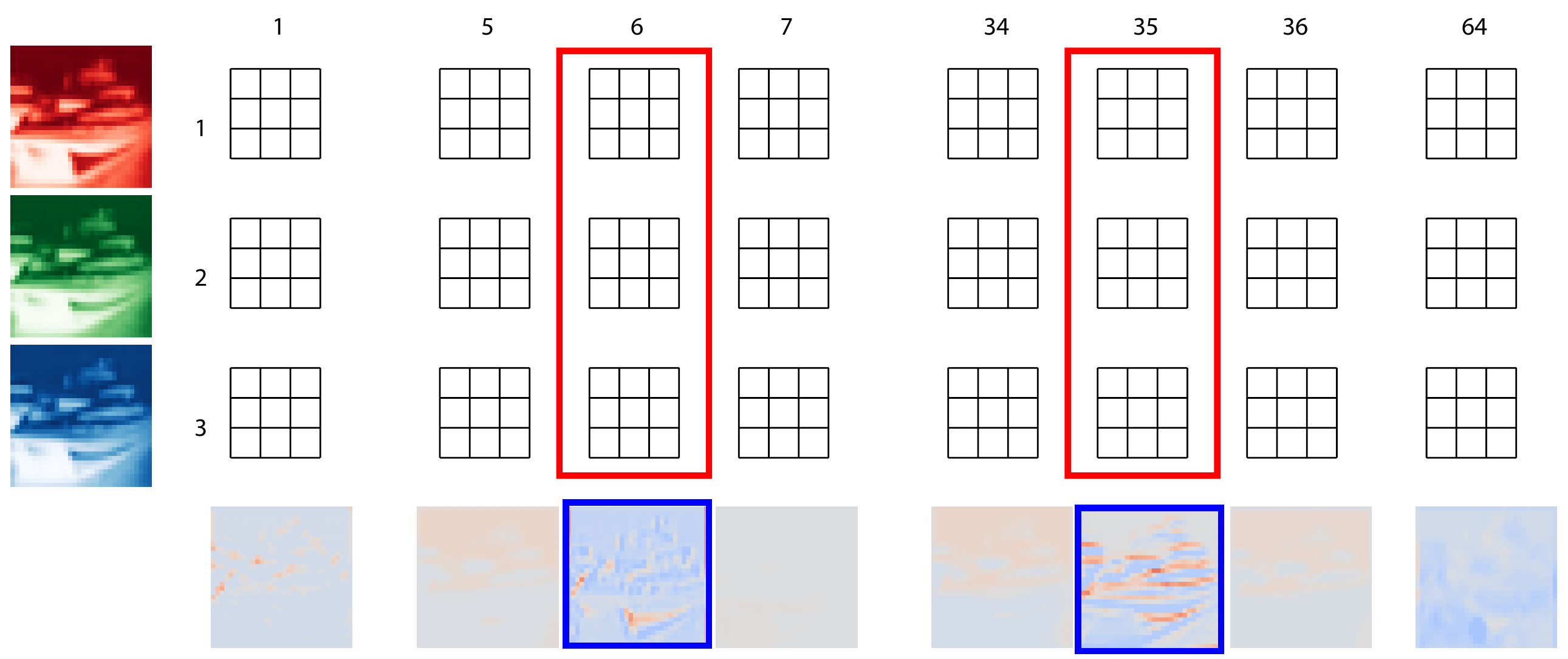}

\end{center}
   \caption{First layer (convolutional layer) of the VGG-16 architecture as an example of parameter layout. In the case of convolutional layer, a \textit{convolutional neuron} is equivalent to a channel, which consists of a set of filters. In the example above, the input contains three channels (R,G,B) and the output contains 64 channels. We name these channels with ordinary numbers from 1 to 64. Due to the space limit, we only show the outputs of channels $1,5,6,7,34, 35, 36, 64$. %
   In this work, we propose to learn the importance of the (output) channels. The two channels outlined in red are the example channels which scored high in the importance. As the output feature maps show (in the blue boxes), the important channels contain humanly-interpretable visual cues. As in structured pruning, we remove the entire channels of less importance such as 7 and 36, while we keep  the informative channels such 6 and 35.}
\label{fig:convlayer}
\end{figure*}

Taken together, we summarize our contributions as follows:
\begin{itemize}{}
\setlength\itemsep{0em}
\item \textbf{A novel pruning technique}. We propose a novel structured pruning technique which learns the importance of the channels for any pre-trained models, providing a practical solution for compressing neural network models. To learn the importance, we introduce an additional, simple operation to the existing neural network architectures, called an \textit{importance switch}. We assigns the Dirichlet distribution over the importance switch, and estimate the parameters of the distribution through variational inference. The learned distribution provides a relative importance of each channel for a task of interest.

\item \textbf{Speedy learning}. Parameter estimation for the importance switch is fast. One epoch is often enough to converge. 



\item \textbf{Insights on neural network compression.} Our method allows us to rank the channels in terms of their learned importance. Visualizing the feature maps of important channels provides insight into which features are essential to the neural network model's task. This intuition explains why neural network compression works at all. 

\item \textbf{Extensive experiments for compression tasks}. We perform extensive experiments to test our method on various architectures and datasets.
By learning which channels are unimportant and pruning them out, our method can effectively compress the networks. Its performance excels across a range of pruning rates.

\end{itemize}


%
%
%
%


\section{RELATED WORK}
\label{sec:relatedwork}
 
The main motivation behind this work is to decrease the size of the network to the set of essential and explainable features, without sacrificing  a model's performance. 
To this end, we slim the network by identifying and removing the redundant channels as a form of structured network pruning \cite{molchanov2017variational, han2015deep}. Compared to weight pruning that removes each individual weight, structured pruning \cite{ifantis1991bounds} that removes channels in convolutional layers or neurons in fully-connected layers, provides practical acceleration. 

Most common pruning approaches take into account the magnitude of the weights and remove the parameters with the smallest L1 or L2-norm  \cite{han2015deep}. Alternatively, gradient information is used to approximate the impact of parameter variation on the loss function \cite{lecun1998gradient, molchanov2017variational}.  In these works, magnitude or a Hessian, respectively, serve as proxies for parameter importance. 

Our work follows the line of research which applies probabilistic thinking to network pruning. A common framework for these methods utilizes Bayesian paradigm and design particular type of priors (e.g. Horseshoe or half-Cauchy prior) which induce sparsity in the network \cite{molchanov2017variational, ullrich2017soft, louizos2017bayesian, oh2019radial}. In our work, we also apply Bayesian formalism, however we do not train the model from scratch using sparse priors. Instead, given any pre-trained model, we learn the importance of the channels and prune out those with less importance, as a post-processing step. We also apply Dirichlet distribution as prior and posterior for learning the channel importance, which has not been seen in the literature. 

Many of the Bayesian approaches assign a distribution over the single weight vector, and, in the case of Bayesian neural networks, perform the variational inference using the mean-field approximation for the computational tractability \cite{balan2015bayesian}, which introduces a large number of parameters, and can be slow or impractical. On the other hand, our approach is practical. It learns the importance of channels as groups of weight vectors, and introduces the number of parameters linear in the number of channels in the network.


One may also find resemblance between the proposed method and attention mechanisms which accentuate certain elements. Dirichlet pruning does something similar, but in a much simpler way. We do not build attention modules (like e.g. \cite{yamamoto2018pcas} which uses neural networks as attention modules), only take a rather simple approach by introducing only the number of Dirichlet parameters equal to the number of channels, and learning them in a Bayesian way.

Dirichlet pruning allows
optimizing single layers at a time, or the entire layers simultaneously as in \cite{yu2018nisp}. In some sense, our work adopts certain aspects of dynamic pruning \cite{gordon2018morphnet} since we automate the neural network architecture design by learning the importance of channels. We perform a short fine-tuning on the remaining channels, resulting in a fast and scalable re-training.

\section{METHOD}
\label{sec:Switches}

Given a pre-trained neural network model, our method consists of two steps.
In the first step, we freeze the original network's parameters, and only learn the importance of the channels (please refer to Fig. \ref{fig:convlayer} for visual definition). In the second step, we discard the channels with low importance, and fine-tune the original network's parameters.
What comes next describes our method in detail.

%


\subsection{Importance switch}

To learn the importance of channels in each layer, we propose to make a slight modification in the existing neural network architecture. We introduce a new component,  \textit{importance switch}, denoted by $\vs_{l}$ for each layer $l$.
Each importance switch is a probability vector of length $D_l$, where $D_l$ is the output dimension of the $l$th fully-connected layer or the number of output channels of the $l$th layer\footnote{Notice that the number of output channels in the layer $l$ is the same as the number of input channels in the layer $l+1$. Importance switch vector $S_l$ is defined over the output channels. However, pruning layer $l$'s output channels also reduces the number of input channels in the layer $l+1$. }. As it is a probability vector, we ensure that the sum across the elements of the vector is 1: $\sum_j^{D_l} \vs_{l,j} = 1$. The switch $\vs_{l,j}$ is the $j$th element of the vector, corresponding to the $j$th output channel on the layer, and its value is learned to represent the normalized importance (as the sum of elements is 1) of that channel.

Introducing a switch operation in each layer in a neural network model may bare similarity to \cite{2017arXiv171201312L, li2019l_0}, where the switch is a binary random variable and hence can only \emph{select} which channels are important. By contrast, our importance switch provides \emph{the degree of importance} of each channel. 

\begin{figure}[t]
\centering
\includegraphics[scale=0.45]{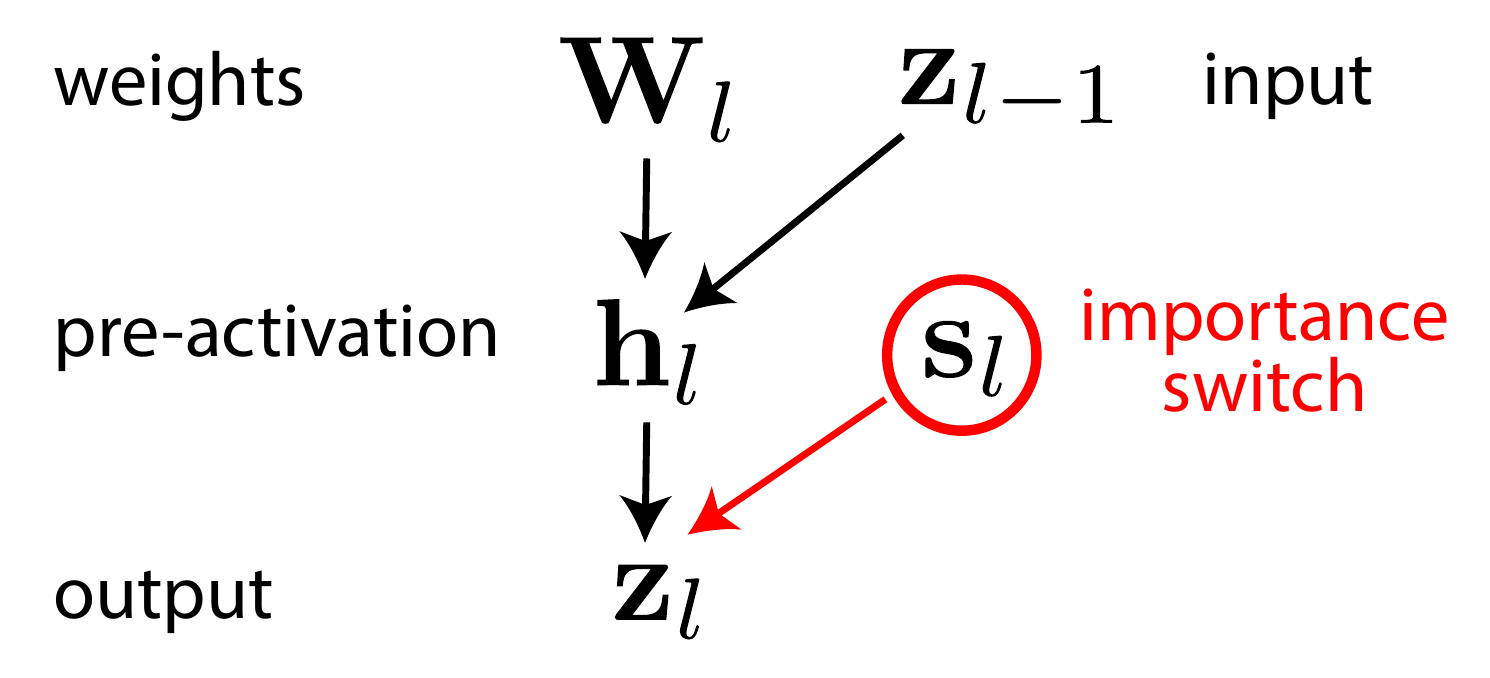}
\caption{Modification of a neural network architecture by introducing \textit{importance switch} per layer. Typically, an input to the $l$th layer $\vz_{l-1}$ and  the weights  $\mathbf{W}_l$ defined by channels form a pre-activation,  which goes through a nonlinearity $\sigma$ to produce the layer's output $\vz_l = \sigma(\vh_l)$. Under our modification, the pre-activation is multiplied by the importance switch then goes through the nonlinearity $\vz_l =\sigma( \mathbf{s}_l \circ \vh_l)$.} \label{fig:schematic_switch}
\end{figure}

With the addition of importance switch, we rewrite the forward pass under a neural network model, where the function $f(\mW_l, \vx_i)$ can be the convolution operation for convolutional layers, or a simple matrix multiplication between the weights $\mW_l$ and the unit $\vx_i$ for fully-connected layers, the pre-activation  is given by  
\begin{align}
    \vh_{l,i} = f(\mW_l, \vx_i),
\end{align} 

%
and the input to the next layer after going through a nonlinearity  $\sigma$, multiplied by a switch $\mathbf{s}_l$, is
\begin{align}
    \vz_{l,i} = \sigma(\mathbf{s}_l \circ \vh_{l,i}),
\end{align} 
where $\circ$ is an element-wise product.

The output class probability under such networks with $L$ hidden layers for solving classification problems can be written as
\begin{align}
    P(\vy_i|\vx_i, \{\mW_{l}\}_{l=1}^{L+1}) &= g \left(\mW_{L+1} \vz_{L,i}\right),
\end{align}
where $\vz_{L,i} = \sigma(\vs_L \circ \left[f(\mW_L \vz_{L-1,i}) \right])$ and $g$ is e.g. the \textit{softmax} operation. 
A schematic of one-layer propagation of the input with the importance switch is given in Fig.~\ref{fig:schematic_switch}. 

\subsection{Prior over importance switch}
We impose a prior distribution over the importance switch using the Dirichlet distribution with parameters $\valpha_0 $:
\begin{align}\label{eq:prior}
    p(\vs_l) &= \mbox{Dir}(\vs_l;\valpha_0).
\end{align} 
Our choice for the Dirichlet distribution is deliberate: as a sample from this Dirichlet distribution sums to 1, each element of the sample can encode  the importance of each channel in that layer.

As we typically do not have prior knowledge on which channels would be more important for the network's output, we treat them all equally important features by setting the same value to each parameter, i.e., $\valpha_0 = \alpha_0*\vone_{D_l}$ where $\vone_{D_l}$ is a vector of ones of length $D_l$ \footnote{Notice that the Dirichlet parameters can take any positive value, $\alpha_i > 0$, however a sample from the Dirichlet distribution is a probability distribution whose values sum to 1}.  When we apply the same parameter to each dimension, this special case of Dirichlet distribution is called \textit{symmetric} Dirichlet distribution. In this case, if we set $\alpha_0<1$ , this puts the probability mass toward a few components, resulting in only a few components that are non-zero, i.e., inducing sparse probability vector. If we set $\alpha_0>1$, all components become similar to each other. Apart from the flexibility of varying $\alpha$, the advantage of Dirichlet probability distribution is that it allows to learn the relative importance which is our objective in creating a ranking of the channels.

\subsection{Posterior over importance switch}

We model the posterior over $\vs_l$ as the Dirichlet distribution as well but with \textit{asymmetric} form to learn a different probability on different elements of the switch (or channels), using a set of  parameters (the parameters for the posterior). We denote the parameters by $\vphi_l$, where each element of the vector can choose any values greater than 0. Our posterior distribution over the importance switch is defined by  
\begin{align}
    q(\vs_l) &= \mbox{Dir}(\vs_l;\vphi_l).
\end{align}

\subsection{Variational learning of importance switches}

Having introduced the formulation of importance switch, we subsequently proceed to describe how to estimate the distribution for the importance switch. 
%
Given the data $\Dat$ and the prior distribution over the importance switch $p(\vs_l)$ given in \eqref{prior}, we shall search for the posterior distribution, 
%
$p(\vs_l|\Dat)$. 
Exact posterior inference under neural network models is not analytically tractable. 
Instead, we resort to the family of \textit{variational} algorithms which attempt to optimize the original distribution $p(\vs_l|\Dat)$ with an approximate distribution $q(\vs)$ by means of minimizing the Kullback-Leibler (KL) divergence:
\begin{equation}
    D_{KL}(q(\vs_l) || (p(\vs_l|\Dat)) 
\end{equation}
which is equivalent to maximizing,
\begin{equation}\label{eq:ELBO}
    \int q(\vs_l) \log p(\Dat|\vs_l) d\vs_l - \mbox{D}_{KL}[ q(\vs_l) || p(\vs_l)],
\end{equation} 
%
%
where
$p(\Dat | \vs_l)$ is the network's output probability given the values of the importance switch. We use \eqref{ELBO} as our optimization objective for optimizing $\vphi_l$ for each layer's importance switch.

Note that we can choose to perform the variational learning of each layer's importance switch sequentially from the input layer to the last layer before the output layer, or the learning of all importance switches jointly (the  details on the difference between the two approaches can be found in the Sec. \ref{sec:experiments}).

During the optimization, computing the gradient of \eqref{ELBO} with respect to $\vphi_l$ requires obtaining the gradients of the integral (the first term) and also the KL divergence term (the second term), as both depend on the value of $\vphi_l$. 
The KL divergence between two Dirichlet distributions can be written in closed form, 
\begin{align*}
    &\mbox{D}_{kl}[ q(\vsbold_l|\vphi_l) || p(\vsbold_l |\valpha_0)] = 
    \log \Gamma (\sum_{j=1}^{D_l} \vphi_{l,j})- \\ & - \log \Gamma (D_l \alpha_{0})
  - \sum_{j=1}^{D_l} \log \Gamma(\vphi_{l,j}) + D_l \log \Gamma(\alpha_{0}) \nonumber \\
    & +  \sum_{j=1}^{D_l} (\vphi_{l,j} - \alpha_{0}) \left[ \psi(\vphi_j) - \psi(\sum_{j=1}^{D_l}\vphi_{l,j} ) \right],
\end{align*} 
where $\vphi_{l,j}$ denotes the $j$th element of vector $\vphi_{l}$, $\Gamma$ is the Gamma function and $\psi$ is the digamma function.
Notice that the first term in \eqref{ELBO} requires broader analysis. 
As described in \cite{NIPS2018_7326}, the usual reparameterization trick, i.e., replacing a probability distribution with an equivalent parameterization of it by using a deterministic and differentiable transformation of some fixed base distribution\footnote{For instance, a Normal distribution for $z$  with parameters of mean $\mu$ and variance $\sigma^2$ can be written equivalently as $z = \mu + \sigma \epsilon$ using a fixed base distribution $\epsilon \sim \Nrm(0,1)$.}, does not work. For instance, in an attempt to find a reparameterization, one could adopt the representation of a $k$-dimensional Dirichlet random variable, $\vs_l \sim \mbox{Dir}(\vs_l|\vphi_l)$, as a weighted sum of Gamma random variables, 
\begin{align}
    \vs_{l,j} &= y_j/(\sum_{j'=1}^K y_{j'}),  \nonumber \\ 
    y_j &\sim \mbox{Gam}(\vphi_{l,j},1) = {y_j^{(\vphi_{l,j}-1)}\exp(-y_j)}/{\Gamma(\vphi_{l,j})}, \nonumber 
\end{align} 
where the shape parameter of Gamma is $\vphi_{l,j}$ and the scale parameter is $1$. However, this does not allow us to detach the randomness from the parameters as the parameter still appears in the Gamma distribution, hence one needs to sample from the posterior every time the variational parameters are updated, which is costly and time-consuming.

\textit{Implicit gradient computation.}
Existing methods suggest either explicitly or \textit{implicitly} computing the gradients of the inverse CDF of the Gamma distribution during training to decrease the variance of the gradients (e.g., \cite{2015arXiv150901631K}, \cite{NIPS2018_7326}, and \cite{pmlr-v80-jankowiak18a}). 
%

\textit{Analytic mean of Dirichlet random variable.}
Another computationally-cheap choice would be using the analytic mean of the Dirichlet random variable to make a point estimate of the integral $\int q_{\vphi_l}(\vs_l) \log p(\Dat|\vs_l) d\vs_l \approx \log p(\Dat|\tilde{\vs}_l) $, where $\tilde{\vs}_{l,j} = \vphi_{l,j}/\sum_{j'=1}^{D_l} \vphi_{l,j'}$, which allows us to directly compute the gradient of the quantity without sampling from the posterior. 

%
%
In our experiments, we examine the quality of posterior distributions learned with computing the gradients of the integral implicitly using the inverse CDF of the Gamma distribution, or with computing the gradients of the integral explicitly using the analytic mean of the Dirichlet random variable, in terms of the quality of learned architectures.

Note that as we add a probability vector (the importance switch) which sums to one, there is an effect of scaling down the activation values. However, once we learn the posterior distribution over the importance  switch, we compress the network accordingly and then retrain the network with the remaining channels to recover to the original activation values.  
Our method is summarized in \algoref{Dirichlet_Pruning}. Also, note that step 3 of \algoref{Dirichlet_Pruning} involves removing unimportant channels. Given the continuous values of posterior parameters, what is the cut-off that decides important channels from the rest at a given layer? In this paper, we search over sub-architectures at different pruning rates, where we select the important channels within those pruning rates as shown in \secref{experiments}. However, other ways, e.g., using the learned posterior uncertainty, can potentially be useful. We leave this as future work.   

\begin{algorithm}[!t]
\caption{Dirichlet Pruning}\label{algo:Dirichlet_Pruning}
\begin{algorithmic}
\vspace{0.1cm}
\REQUIRE A pre-trained model, $\mathcal{M}_\theta$ (parameters are denoted by $\theta$).
\vspace{0.1cm}
\ENSURE Compressed model $\hat{\mathcal{M}}_{\hat\theta}$ (reduced parameters are denoted by ${\hat\theta})$.  \\
\STATE \textbf{Step 1}. Add importance switches per layer to $\mathcal{M}_\theta$.
\STATE  \textbf{Step 2}.
Learn the importance switches via optimizing \eqref{ELBO}, with freezing $\theta$.
\STATE  \textbf{Step 3}. Remove unimportant channels according to the learned importance. 
\STATE  \textbf{Step 4}. Re-train $\hat{\mathcal{M}}_{\hat\theta}$ with remaining channels.
\end{algorithmic}
\end{algorithm}


\section{EXPERIMENTS}
\label{sec:experiments}

In this section we apply the proposed method to create pruned architectures. The compression rates have been evaluated against a variety of existing common and state-of-the-art benchmarks, with the focus on probabilistic methods. We then also 
demonstrate how the important channels selected by our method may contain (human-perceivable) distinct visual features. 
The experiments are performed on three datasets, MNIST and FashionMNIST, which are used to train the LeNet-5 network, and CIFAR-10 used to train the ResNet-56, WideResNet-28-10 and VGG-16.



\subsection{Variants of Dirichlet pruning} Dirichlet pruning is a flexible solution which allows for several variants. In the implementation of the importance switch parameter vector, the posterior distribution over switch via the variational inference objective as given in ~\eqref{ELBO} is evaluated. To compute the gradients of the integral (cross-entropy term) implicitly we use the samples from the inverse CDF of the Gamma distribution. For a given layer with $n$ output channels we draw $k$ samples of the importance switch vectors of length $n$. For Lenet-5 network we sample for $k = 50, 150, 300, 500$ and for VGG16 we sample for $k=10, 20, 50, 100$ (the number of samples are provided in brackets when needed, e.g Dirichlet (300)).

In addition, we include the variant of the method where we compute the gradients of the integral explicitly using the analytic mean of the Dirichlet random variable (in the supplementary materials, we include an additional toy experiment which tests the difference between the two approaches). In the above approaches, we compute the importance switch vector for each layer separately. However, we are also able to train switch values for all the layers in one common training instance. This case is denoted by ``joint"  in brackets, e.g., Dirichlet (joint).
%
%

  \begin{table}[t!]
    \centering
    
     \begin{tabular}{ccccc}
    \toprule

        Method & Error & FLOPs & Params\\
  	\midrule
          \textbf{Dirichlet (150)} & 1.1 & 168K & \textbf{6K} \\ 
          
          \textbf{Dirichlet (mean)} & 1.1 & 140K & \textbf{5.5K} \\ 
          
          \textbf{Dirichlet (joint)} &1.1 & 158K & \textbf{5.5K} \\ 
        
          BC-GNJ \cite{louizos2017bayesian} & 1.0 & 288K & 15K\\ 
          BC-GHS \cite{louizos2017bayesian} & 1.0 & 159K & 9K\\ 
          
          RDP \cite{oh2019radial} & 1.0 & 117K & 16K\\  
          FDOO (100K) \cite{tang2018flops} & 1.1 &  \textbf{113K} & 63K \\ 
          FDOO (200K) \cite{tang2018flops} & 1.0 & 157K & 76K\\ 
           GL \cite{wen2016learning} &  1.0 & 211K & 112K\\ 
           GD \cite{srinivas2015data} &  1.1 & 273K & 29K\\ 
            SBP \cite{neklyudov2017structured} &  \textbf{0.9} & 226K & 99K \\
        \bottomrule


          
          
        
          

    \end{tabular}
    \caption{The structured pruning of \textbf{LeNet-5}. The pruned network is measured in terms of the number of FLOPs and the number of parameters (Params).
    The proposed method outperforms the benchmark methods as far as the number of parameters is concerned and it produces the most optimal Params to FLOPs ratio. 
    }
    
    \label{tab:compression_rank}

\end{table}


When computing the importance switch, we load the pretrained model in the first phase, and then add the importance switch as new parameters. We then fix all the other network parameters to the pretrained values and finetune the extended model to learn the importance switch. In compression process, we mask the subsets of features (both weights and biases, and the batch normalization parameters). 

\subsection{Compression}

The goal of the neural network compression is to decrease the size of the network in such a way that the slimmer network which is a subset of the larger network retains the original performance but is smaller (which is counted in network parameters) and faster (counted in floating points operations or FLOPs). The bulk of the parameter load comes from the fully-connected layers and most of the computations are due to convolutional operations, and therefore one may consider different architectures for different goals.

We tackle the issue of compression by means of the Dirichlet pruning method in a way that the network learns the probability vector over the channels, that is where the support of the distribution is the number of channels. The channels that are given higher probability over the course of the training are considered more useful, and vice-versa. The probabilities over the 



  \begin{table}[t]
  \begin{tabular}{cccc}
  \toprule
Method & Error & FLOPs & Parameters \\
\midrule
\textbf{Dirichlet (ours)} & \textbf{8.63} & \textbf{46.0M} & \textbf{0.73M} \\
Hrank \cite{lin2020hrank} & 8.77 & 73.7M  & 1.78M  \\
BC-GNJ \cite{louizos2017bayesian} & 8.3 & 142M &  1.0M \\
BC-GHS \cite{louizos2017bayesian} & 9.0 & 122M & 0.8M  \\
RDP \cite{oh2019radial} & 8.7 & 172M & 3.1M  \\
GAL-0.05 \cite{lin2019towards} & 7.97 & 189.5M  & 3.36M  \\
SSS \cite{huang2018data} & 6.98 & 183.1M  & 3.93M  \\
VP \cite{zhao2019variational} & 5.72 & 190M & 3.92M  \\
\bottomrule
\end{tabular} 
\caption{\textbf{VGG-16} on CIFAR-10. Dirichlet pruning produces significantly smaller and faster models.}
\label{tab:compression_rank_vgg}
\end{table}

    \begin{table} [t]
  \begin{tabular}{cccc}
  \toprule
Method & Error & FLOPs & Parameters \\
\midrule
\textbf{Dirichlet (ours)} & \textbf{8.83} & 45.64M & \textbf{0.26M} \\
Hrank \cite{lin2020hrank} & 9.28 & \textbf{32.53M} & 0.27M\\
GAL-0.8 \cite{lin2019towards} & 9.64 & 49.99M & 0.29M \\
CP \cite{he2017channel} & 9.20 & 62M & - \\
\bottomrule
\end{tabular} 
\caption{\textbf{ResNet-56} on CIFAR-10.  Our method outperforms the recent methods, in particular when it comes to the model size (benchmark results come from the original sources). In the ResNet implementation, we use the approximation using the analytic mean. }
\label{tab:compression_rank_resnet}
\vspace{-0.2cm}
\end{table}
 \begin{table} [H]
    \centering
     \begin{tabular}{cccc}
    \midrule
        Method & Error & Comp. Rate & Params\\
        \hline
         \textbf{Dirichlet (ours) }   & 
         4.5 & 52.2\% 
         & \textbf{17.4M}\\
         
        $L_0$ ARM \cite{li2019l_0}  & 
        4.4 & $49.9\%$
        & 18.3M\\
        
        $L_0$ ARM \cite{li2019l_0} & \textbf{4.3} & $49.6\%$ & 18.4M\\
          \bottomrule
    \end{tabular}
    \caption{\textbf{WideResNet-28-10} on CIFAR-10. Compared to $L_0$-ARM, with a slight increase in the error rate, our method achieves the smallest number of parameters. 
    }
    \label{tab:compression_rank_wrn}
    \vspace{-0.3cm}
\end{table}

channels can be ordered, and the channels which are given low probability can be pruned away. Subsequent to pruning, we retrain the network on the remaining channels. 

In the case of LeNet and VGG networks, we consider all the channels in every layer. In the case of residual networks each residual block consists of two convolutional layers. To preserve skip connection dimensionality in a similar fashion to \cite{li2019l_0}, we prune the output channels of the first convolutional layer (equivalently input channels to the second layer). ResNet-56 consists of three sections with all convolutional layers having 16, 32 and 64 channels, respectively. Similarly, WideResNet-28-3 has 12 residual blocks (three sections of four blocks with 160, 320, 640 channels, respectively). We fix the number of channels pruned for each section. A finer approach could further bring better results.

 \begin{figure}[H]
  \centering
  \subfloat[]{\includegraphics[scale=0.35]{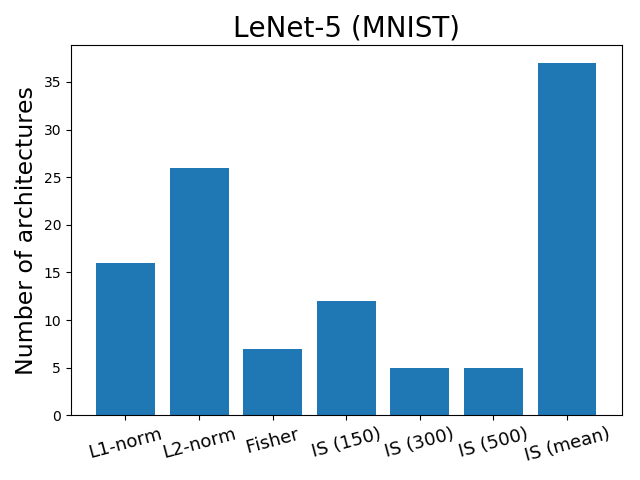}\label{fig:2}}
  \vspace{1em}
  \subfloat[]{\includegraphics[scale=0.35]{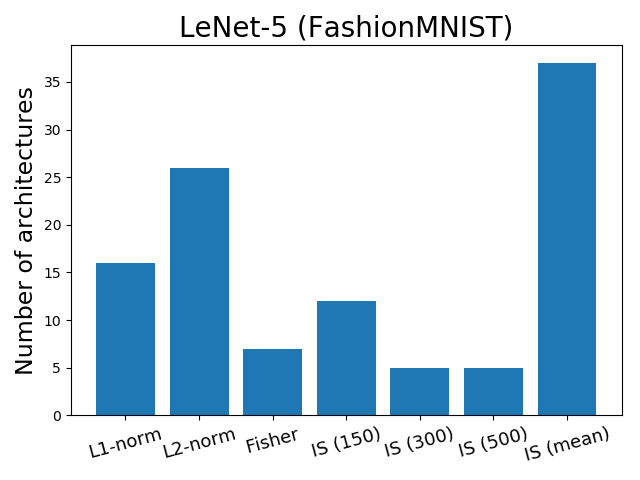}\label{fig:3}}
   \vspace{1em}
   \subfloat[]{\includegraphics[scale=0.35]{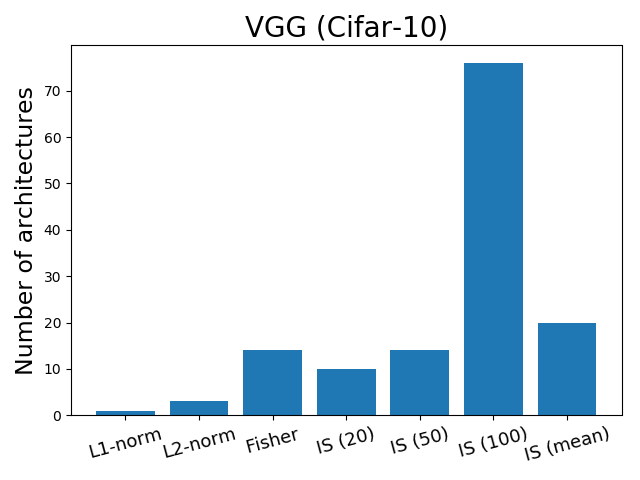}\label{fig:1}}
  \caption{Frequencies of best sub-architectures selected by each method. 
  Considering 108 sub-architectures for LeNet-5 and 128 sub-architectures for VGG,  the height of each bar describes the number of sub-architectures pruned by each method where a given method achieved the best test performance.
  %
  We compare seven methods, including four variants of Dirichlet pruning, which we label by importance switch (IS). In all cases, our method dominantly performs over the largest set of sub-architectures, suggesting that the performance of our method is statistically significant.
  }
  \label{fig:bar_chart}
  \vspace{-0.4cm}
\end{figure}

\begin{figure*}[th!]
    \centering
    \subfloat[\textbf{MNIST}]{{\includegraphics[width=0.47\linewidth]{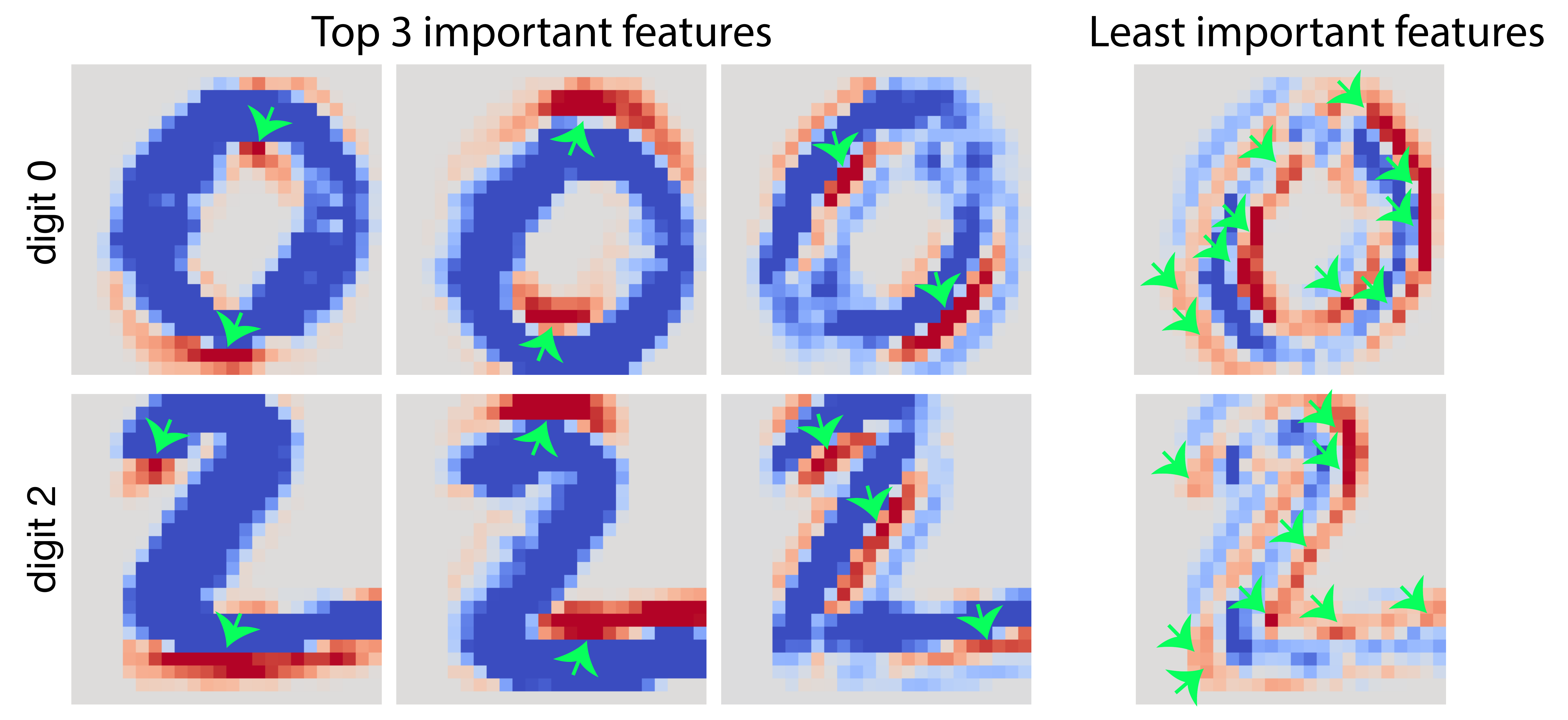} }}%
    \vspace{1em}
    \subfloat[\textbf{FashionMNIST}]{\includegraphics[width=0.48\linewidth]{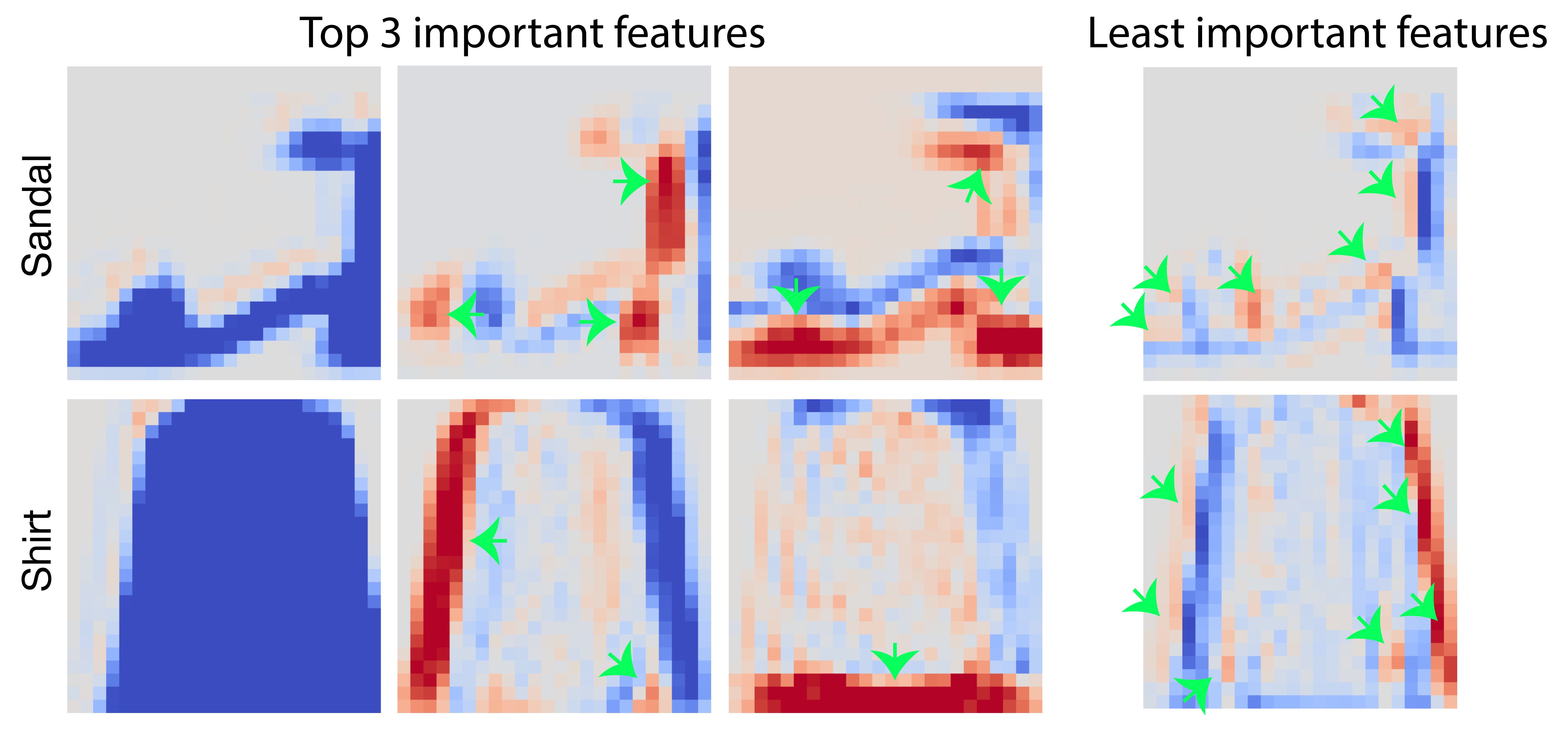}}%
    \caption{Visualization of learned features for two examples from MNIST and FashionMNIST data for top three (the most important) features and bottom one (the least important) feature. Green arrows indicate where high activations incur. The top, most significant features exhibit strong activations in only a few  \textit{class-distinguishing} places in the pixel space. Also, these features exhibit the complementary nature, i.e., the activated areas in the pixel space do not overlap among the top 3 important features. 
    On the other hand, the bottom, least significant features are more fainter and more scattered. 
}%
    \label{fig:example}%
    \vspace{0.5cm}
\end{figure*}

\subsubsection{Compression rate comparison.}

Table~\ref{tab:compression_rank} presents the results of LeNet trained on MNIST, Table~\ref{tab:compression_rank_vgg} the results of VGG trained on CIFAR-10. Moreover, we test two residual networks with skip connections,  Table~\ref{tab:compression_rank_resnet} 
includes the results of ResNet-56 and
Table~\ref{tab:compression_rank_wrn} demonstrates the results on WideResNet-28-10, both also trained on CIFAR-10. In the first test, we compare the results against the existing compression techniques, several of which are state-of-the-art Bayesian methods (we adopt the numbers from each of the papers). In the next subsection given the available codebase, we perform a more extensive search with magnitude pruning and derivative-based methods.

 Note that our proposed ranking method produces very competitive compressed architectures,  producing smaller (in terms of parameters) and faster (in terms of FLOPs) architectures with the similar error rates. In particular for LeNet, the compressed architecture has 5.5K parameters which is less than all the other methods, and 140K FLOPs which is third to RDP and FDOO(100K) that, however, have over three and ten times more parameters, respectively. The method works especially well on VGG producing an architecture which is smaller than others in the earlier layers but larger in later layers. This effectively reduces the number of required FLOPs compared to other state-of-the-art methods (44M in our case, two times less compared the second, HRank) for similar accuracy. The proposed methods are general and work for both convolutional and fully-connected layers, however they empirically show better results for convolutional layers. We believe that this behavior comes from the fact that these channels consist of a larger number of parameters and therefore are less affected by noise during SGD-based training (which gets averaged over these parameters), and therefore their importance can be measured more reliably.

\subsubsection{Search over sub-architectures}

In the second experiment for each method we verify method's pruning performance on a number of sub-architectures. 
We design a pool of sub-architectures with a compression rate ranging 20-60\%. As mentioned earlier, some of the practical applications may require architectures with fewer convolutional layers to cut down the time and some may just need a network with smaller size. For Lenet-5 we use 108 different architectures and for VGG we test 128 architectures. We use the most popular benchmarks whose code is readily available and can produce ranking relatively fast. These are common magnitude benchmarks, L1- and L2-norms and the state-of-the art second derivative method based on Fisher pruning \cite{crowley2018closer, theis2018faster}. 
\figref{bar_chart} shows the number of times each method achieves superior results to the others after pruning it to a given sub-architecture. Dirichlet pruning works very well, in particular, for the VGG16 among over 80\% of the $128$ sub-architectures we considered, our method achieves better accuracy than others.

\subsection{Interpretability} 
\label{sub: interpretability}

In the previous sections we describe the channels numerically. In this section, we attempt to characterize them in terms of visual cues which are more human interpretable. In CNNs, channels correspond to a set of convolutional filters which produce activations that can be visualized \cite{yosinski2015understanding, mahendran2016visualizing}. 
Visualization of the first layer's feature maps provides some insight into how the proposed method makes its decisions on selecting important channels. As we presented the example from CIFAR-10 in \figref{convlayer}, the feature maps of the important channels  contain stronger signals and features that allow humans to identify the object in the image. In contrast, the less important channels contain features which can be less clear and visually interpretable. 

In \figref{example},
we visualize feature maps produced by the first convolution layer of LeNet network given two example images from the MNIST and Fashion-MNIST, respectively. In contrast to the VGG network, almost all feature maps in LeNet allow to recognize the digit of the object. However, the important features tend to better capture distinguishing features, such as shapes and object-specific contour.
In the MNIST digits, the learned filters identify \textit{local parts} of the image (such as lower and upper parts of the digit '2' and opposite parts of the digit '0'). 
On the other hand, the most important feature in the  FashionMNIST data is the overall shape of the object in each image, that is each class has different overall shape (e.g., shoes differ from T-shirts, bags differ from dresses).

The visualization of first layer's feature maps produced by the important channels helps us to understand why the compressed networks can still maintain a similar performance as the original immense networks. This seems to be because the compressed networks contain the core class-distinguishing features, which helps them to still perform a reliable classification even if the models are now significantly smaller.  That being said, interpretability is a highly undiscovered topic in the compression literature.  The provided examples illustrate the potential for interpretable results but a more rigorous approach is a future research direction.

\section{Conclusion}

Dirichlet pruning allows compressing any pre-trained model by extending it with a new, simple operation called \textit{importance switch}. To prune the network, we learn and take advantage of the properties of Dirichlet distribution.  Our choice for the Dirichlet distribution is deliberate. (a) A sample from Dirichlet distribution is a probability vector which sums to 1. (b) Careful choice of Dirichlet prior can encourage the sparsity of the network. (c) Efficient Bayesian optimization thanks to the closed-form expression of the KL-divergence between Dirichlet distributions.
Thus, learning Dirichlet distribution allows to rank channels according to their relative importance, and prune out those with less significance. 
Due to its quick learning process and scalability, the method works particularly well with large networks, producing much slimmer and faster models. 
Knowing the important channels allows to ponder over what features the network deems useful.
An interesting insight we gain through this work is that
the features which are important for CNNs are often also the key features which humans use to distinguish objects.

\clearpage

\section*{Acknowledgments}

The authors are supported by the Max Planck Society.
Mijung Park is also supported by the Gibs Sch{\"u}le Foundation and the Institutional Strategy of the University of T{\"u}bingen (ZUK63) and the German Federal Ministry of Education and Research (BMBF): T\"ubingen AI Center, FKZ: 01IS18039B.
Kamil Adamczewski is grateful for the support of the Max Planck ETH
Center for Learning Systems.

\section*{Code}

The most recent version of the code can be found at \url{https://github.com/kamadforge/dirichlet_pruning}. The stable version for reproducibility can also be found at \url{https://github.com/ParkLabML/Dirichlet_Pruning}.

\section*{REFERENCES}

\bibliographystyle{plain}
\bibliography{Main}

\end{document}


\maketitle

\setcounter{section}{0}

\section{Comparison between the posterior distributions learned by taking implicit gradients using the inverse CDF of Gamma and by taking explicit gradients using the analytic mean of Dirichlet.  }\label{supp:Compare_sampling_nonsampling}



 We consider a model with a single hidden-layer for a binary classification problem, with varying dimensions for input and hidden-layer, where input dimension $d_x$ and hidden dimension $d_h$ pairs we used are $(d_x =100,d_h=20)$; $(d_x=500,d_h=200)$; and $(d_x=1000, d_h=500)$.  We assume that there is a known set of ground-truth switch values (which we simulated) in each of these models, as well as a known set of ground-truth  parameter values $W_1, W_2, b_1, b_2$ (for one hidden layer MLP, which we also simulated). Then, we generate inputs from an isotropic Gaussian with zero vector mean and variance 1, which we propagate through each of the models for producing the class labels 0. We also generate inputs from an isotropic Gaussian with the vector of 2's for the mean and variance 0.2 and generated the class labels 1.  
Given these datasets, under each model, we then compute the posterior distributions over the importance switch.


\begin{figure}[h]

\begin{center}
  \includegraphics[width=0.7\linewidth]{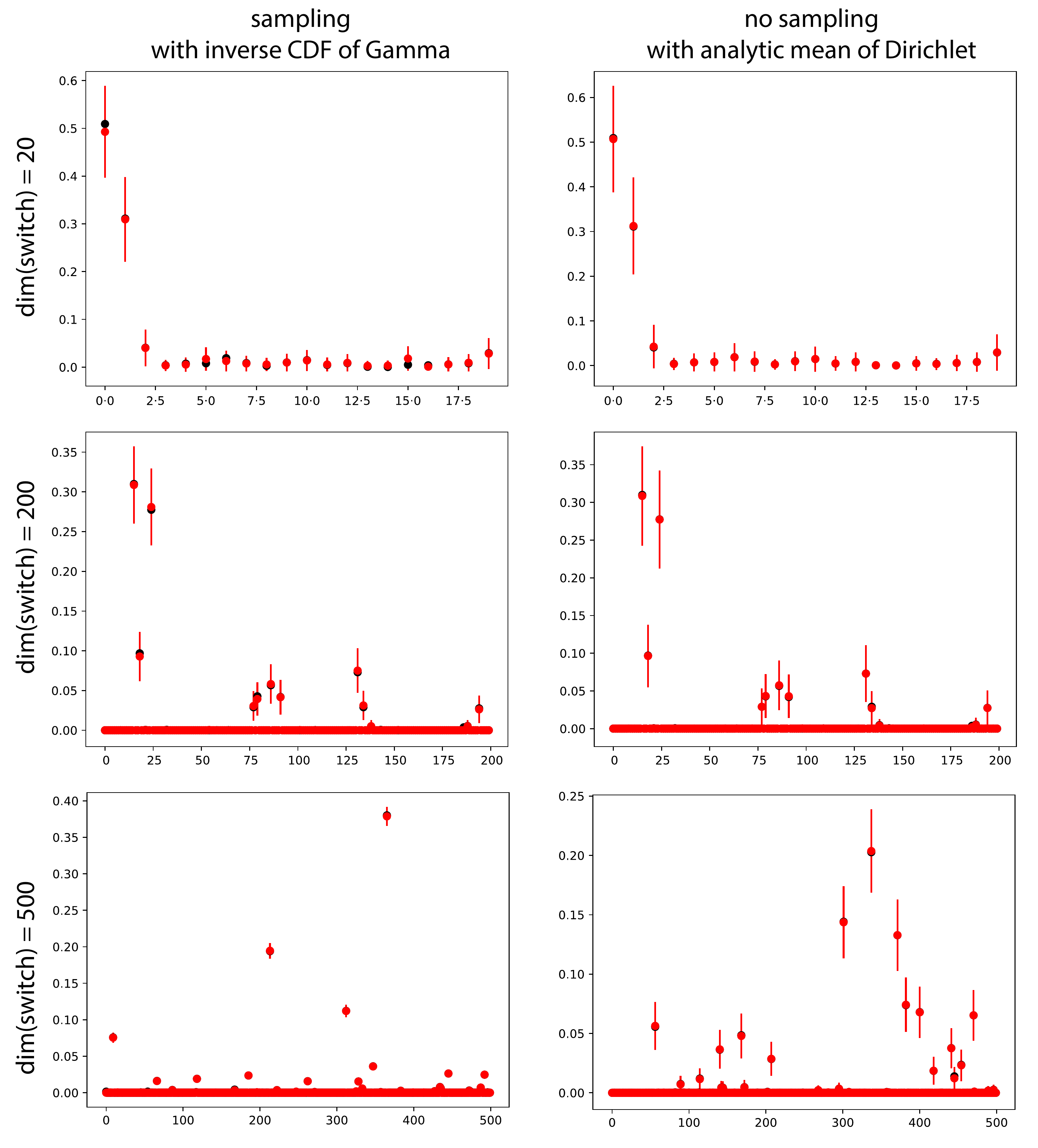}
  \caption{Comparison of posterior distributions with varying switch dimension. Black dots indicate true switch values from which we generated data. Red dots indicate the posterior means and the red vertical lines are one posterior standard deviation. }
\end{center}

\end{figure}\label{fig:Compare}

In \figref{Compare},  we find that the posterior distribution (Left column) with the inverse CDF of the Gamma distribution provides smaller posterior variance (more certain) than the one with analytic mean, regardless of the dimension of the switch, while the difference in variance gets larger with a higher dimension of the switch. However, the difference in the posterior means in two cases is negligible. In terms of the computation time, using the analytic mean of Dirichlet is a clear winner (no need to sample from the posterior), significantly faster than the case with sampling and using the inverse CDF of the Gamma distribution for implicit gradient computation. 
For instance, when $d_x=1000$ and $d_h = 500$, drawing $1000$ samples from the Gamma distribution and computing the Monte Carlo integration of the integral, and computing the gradients of the integral wrt the parameters of the switch took\footnote{This computation time is measured on a desktop using Intel Xeon W-2135 CPUs at 3.70GHz.} $14.317$ seconds for performing one epoch training with mini-batch size $100$ for the dataset size $N=4000$. On the other hand, without drawing samples, simply taking the analytic mean of the Dirichlet to approximate the integral and computing the gradients of this approximation with respect to the parameters of the switch takes only $0.570$ seconds for the same one epoch training. 

\section{Descriptions of the benchmark  methods.}\label{supp:benchmark_method}

In the case of the compression experiment, the following benchmark methods are used:

\begin{itemize}
\setlength\itemsep{1pt}

\item L1-norm and L2-norm. In the case of convolutional neurons we compute L1- or L2-norm of the parameters which belong to the given channel $n_i$ for $n_i \in [1,2,\dots, N_l]$. In the case of fully-connected neuron for $n_i \in [1,2,\dots, N_l]$, we compute L1- or L2-norm of all the incoming weights from the previous layer to the parameter $n_i$

\item Group Lasso (GL) \cite{wen2016learning} - the method combines non-structured regularization applied to every weight with structured sparsity regularization applied to every layer. Then for groups of parameters it uses group regularization during training using L2-norm (the name of the method is a misnomer).

\item Derivative-based pruning \cite{lecun1998gradient, theis2018faster}, where we compute the importance of the channel based on the gradient of the following activation. $\mathbb{R}^{H_l \times W_l \times C_l} \xrightarrow{} \mathbb{R}$. Let $h_i$ be the output produced by parameter (or in our case channel or a set of parameters) and $\mathcal{C}$ be a cost function, negative log-likelihood being the most common choice. Then following \cite{molchanov2016pruning} the cost of removing this channel is $$\abs{\Delta \mathcal{C}(h_i=0)} =\abs{\diffp{\mathcal{C}}{{h_{i}}} h_i}.$$

\item Bayesian Compression for Deep Learning (BD) \cite{louizos2017bayesian} - the approach proposes to phrase the compression problem as a variational inference optimization with hierarchical sparsity-inducing prior to prune groups (G) of parameters (nodes) in a CNN. The first method (BC-GNJ) uses the improper log-uniform prior and its alternative reparametrization known as normal-Jeffrey (NJ) prior. The second method (BC-GHS) incorporates prior hierarchy that uses two Half-Cauchy  distributions, which induce horseshoe (HS) prior distribution of the weights. 

\item Flops as a Direct Optization Objective (FDOO) \cite{tang2018flops} - the method concentrates on the problem of reducing FLOPs (rather than parameters). They construct a loss function which incorporates the number of FLOPs as a variable. The loss function is phrased as a fully-factorized spike-and-slab posterior with the number of FLOPs as a sparsity inducing prior. The optimization of the FLOPS uses score function estimator (REINFORCE). Two results correspond to the best compression rates given two FLOPs budgets (100K and 200K FLOPs).

\item Structured Bayesian Pruning via Multiplicative Noise (SBP) \cite{neklyudov2017structured} - the authors introduce a dropout-like layer with a certain kind of multiplicative noise (Bernoulli or Gaussian). Besides, sparsity-inducing log-uniform prior is used, but it is placed over the noise variables rather than weights. The sparsification is obtained during the variational lower bound training.

\item Data-free parameter pruning \cite{srinivas2015data} - the method uses neuron similarity to remove redundant neurons. Assuming the same activation for all the outputs, the neurons which have similar weights are first combined and then discarded.

\end{itemize}{}

\section{The layer-by-layer compressed architectures.}

\begin{table}[h]
    \centering
    \begin{tabular}{ccccc}
    \toprule

        Method & Architecture & Error & FLOPs & Params\\
  	\midrule
          \textbf{Dirichlet (150)} &\textbf{6-8-40-20}&  1.1 & 168K & \textbf{6K} \\ 
          
            \textbf{Dirichlet (mean)} &\textbf{5-8-45-15}&  1.1 & 140K & \textbf{5.5K} \\ 
          
          \textbf{Dirichlet (joint)} &\textbf{6-7-35-17}&  1.1 & 158K & \textbf{5.5K} \\ 
        
          BC-GNJ \cite{louizos2017bayesian} & 8-13-88-13&  1.0 & 288K & 15K\\ 
          BC-GHS \cite{louizos2017bayesian} & 5-10-76-16 & 1.0 & 159K & 9K\\ 
          
          RDP \cite{oh2019radial} & 4-7-110-66 & 1.0 & 117K & 16K\\  
          FDOO (100K) \cite{tang2018flops} & 2-7-112-478&  1.1 &  \textbf{113K} & 63K \\ 
          FDOO (200K) \cite{tang2018flops} & 3-8-128-499& 1.0 & 157K & 76K\\ 
           GL \cite{wen2016learning} & 3-12-192-500&  1.0 & 211K & 112K\\ 
           GD \cite{srinivas2015data} & 7-13-208-16&  1.1 & 273K & 29K\\ 
            SBP \cite{neklyudov2017structured} & 3-18-284-283& \textbf{0.9} & 226K & 99K \\
        \bottomrule
          
    \end{tabular}
    
        \caption{The structured pruning of \textbf{LeNet-5}.}
    
\end{table}


\begin{table}[h]
    \centering
     \begin{tabular}{ccccc}
    \midrule
        Method & Architecture & Error & FLOPs & Params\\
        \hline
         \textbf{Dirichlet (50)} & \textbf{39-39-63-48-55-88-87-52-62-22-42-47-47-47}   & 8.6 & \bf{46M} & \bf{0.7M}\\

        
        \textbf{Dirichlet (mean)} & \textbf{34-34-68-68-75-106-101-92-102-92-92-67-67-62-62}  & 8.6 & \bf{46M} & \bf{0.7M}\\
        
        
        
        
        
        
         BC-GNJ & 63-64-128-128-245-155-63-26  -24-24-20-14-12-11-15&  8.3 & 142M & 1.0M\\

          BC-GHS & 51-62-125-128-228-129-38-13-9-6-5-6-6-20 & 9.0 & 122M & 0.8M\\

        
        RDP  & 27-57-125-122-236-244-246-340-127-77-89-52-380-414&  8.7  & 172M & 3.1M \\ 
              
          \bottomrule
    
    \end{tabular}
    \caption{The structured pruning of  \textbf{VGG-16}.
    }
    
    \label{tab:compression_rank_vgg}

\end{table}


\begin{table}[h]
    \centering
     \begin{tabular}{ccccc}
    \midrule
        Method & Architecture & Error & Comp. Rate & Params\\
        \hline
        
         \textbf{Dirichlet } & \textbf{65,75,70,90,149,149,164,149,305,325,304,306}   & 
         4.5 & 52.2\% 
         & \textbf{17.4M}\\
        
        $L_0$ ARM \cite{li2019l_0} & \text{82-75-82-87-164-169-156-161-317-317-317-324}  & 
        4.4 & $49.9\%$
        & 18.3M\\
        
        $L_0$ ARM \cite{li2019l_0} & 75-72-78-78-157-165-131-162-336-325-331-343 & \textbf{4.3} & $49.6\%$ & 18.4M\\

          \bottomrule
    
    \end{tabular}
    \caption{The structured pruning of \textbf{WideResNet-28-10}. 
    }
    
    \label{tab:compression_rank_wrn}

\end{table}

\newpage

\section{The complete channel visualization for the VGG-16 first convolutional layer trained on CIFAR-10.}

\begin{center}

\begin{figure}[h!]
\centering
\begin{tabular}{cccccccc}
\includegraphics[width = 1.4cm]{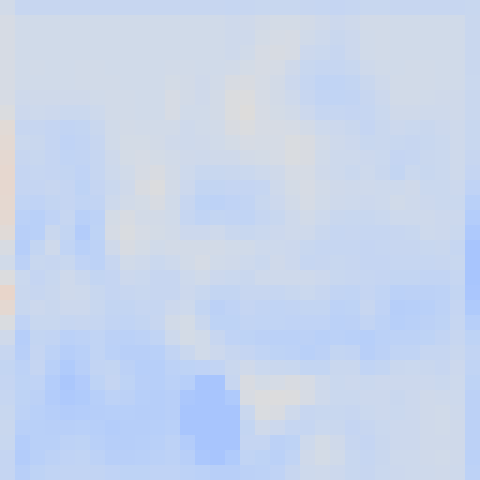} & 
\includegraphics[width = 1.4cm]{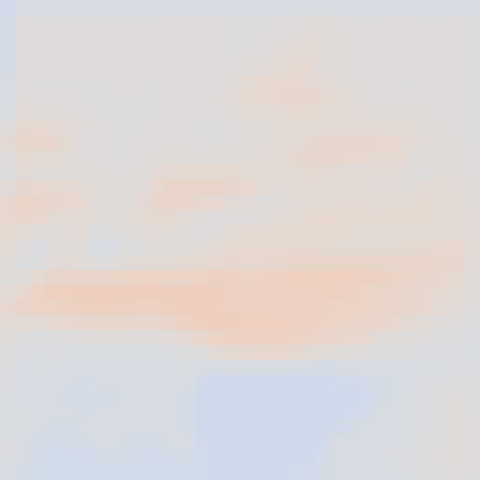} & 
\includegraphics[width = 1.4cm]{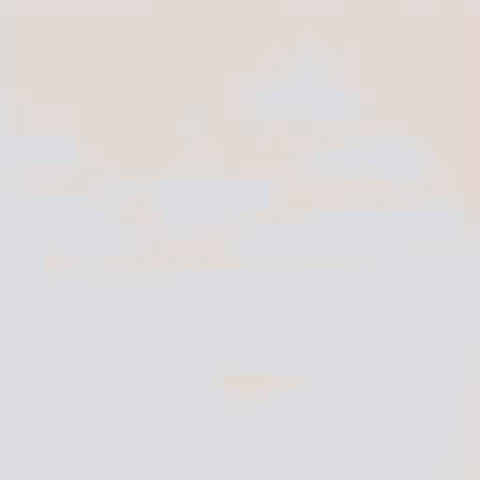} & 
\includegraphics[width = 1.4cm]{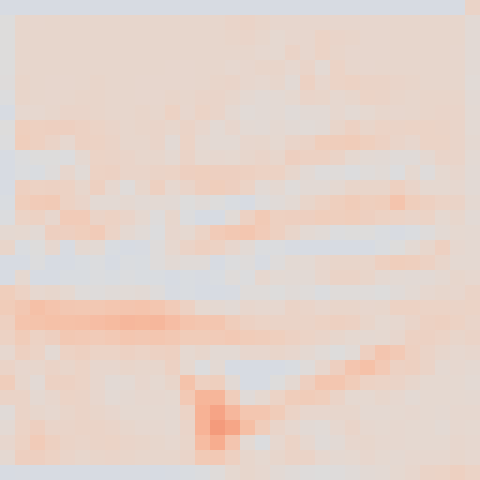} & 
\includegraphics[width = 1.4cm]{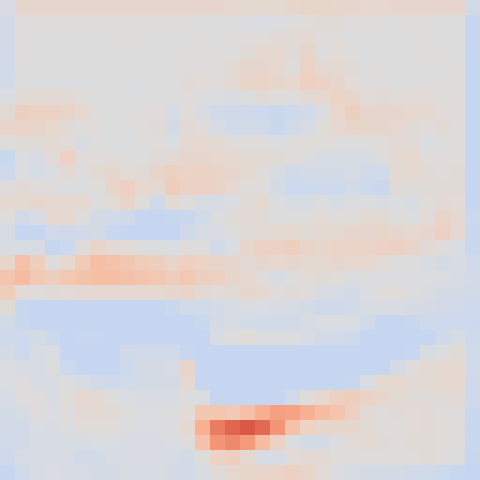} & 
\includegraphics[width = 1.4cm]{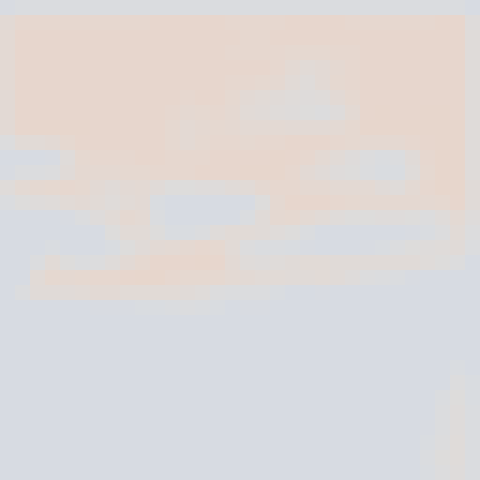} & 
\includegraphics[width = 1.4cm]{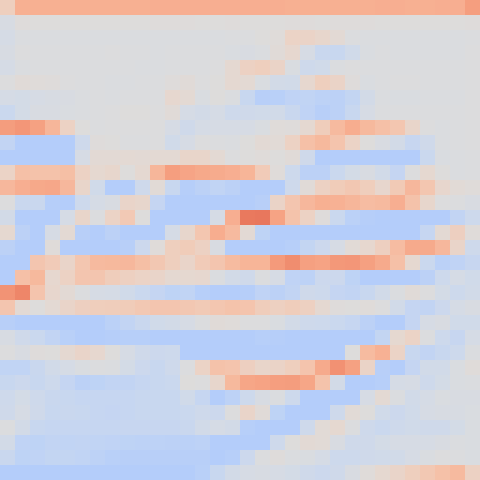} & 
\includegraphics[width = 1.4cm]{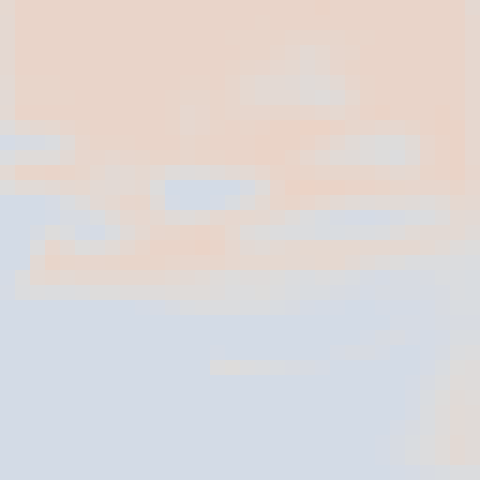}\\
0 & 1 & 2 & 3 & 4 & 5 & 6 & 7  \\
\includegraphics[width = 1.4cm]{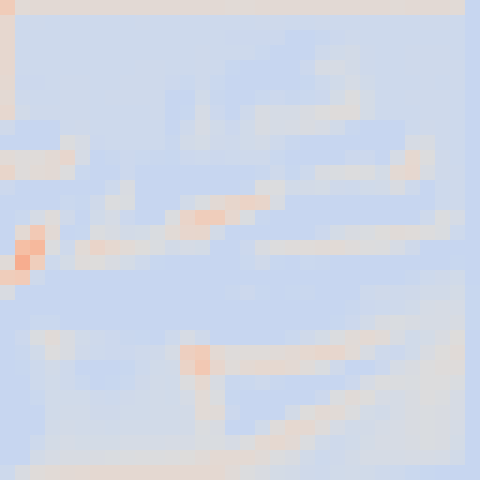} & 
\includegraphics[width = 1.4cm]{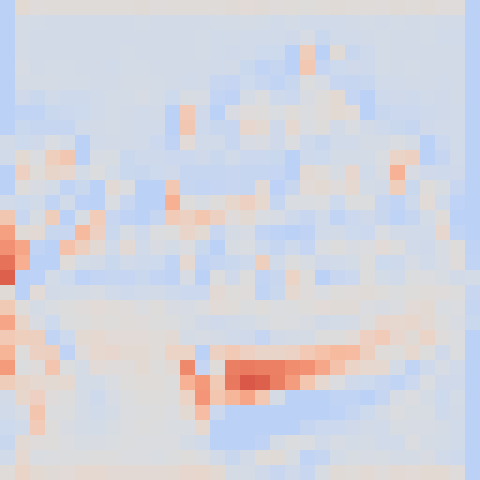} & 
\includegraphics[width = 1.4cm]{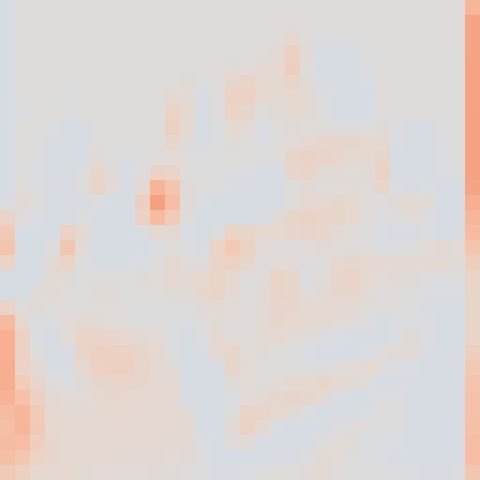} & 
\includegraphics[width = 1.4cm]{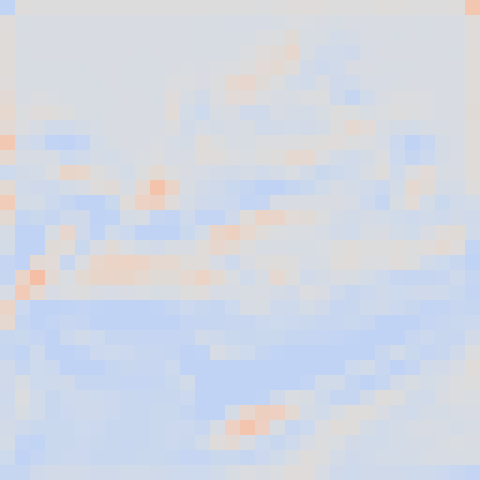} & 
\includegraphics[width = 1.4cm]{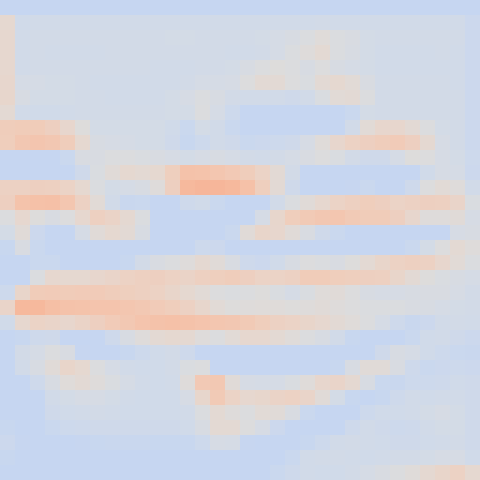} & 
\includegraphics[width = 1.4cm]{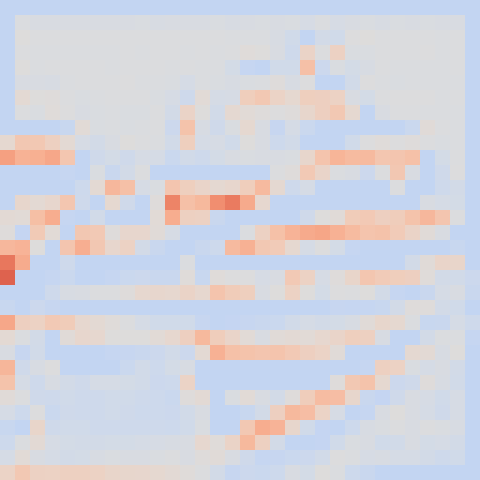} & 
\includegraphics[width = 1.4cm]{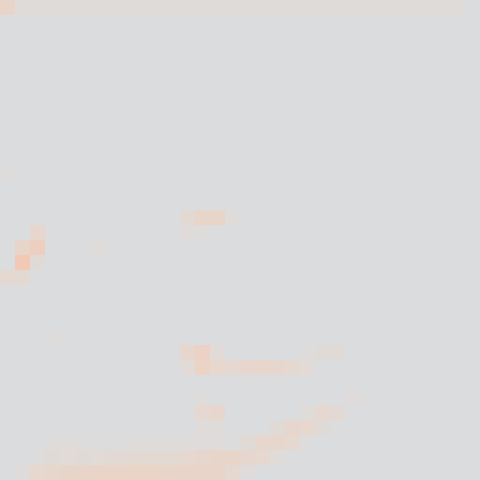} & 
\includegraphics[width = 1.4cm]{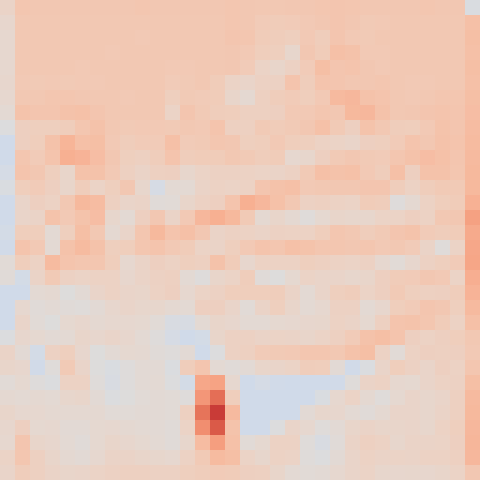} & 

8& 9 & 10& 11 & 12 & 13 & 14 & 15  \\

\includegraphics[width = 1.4cm]{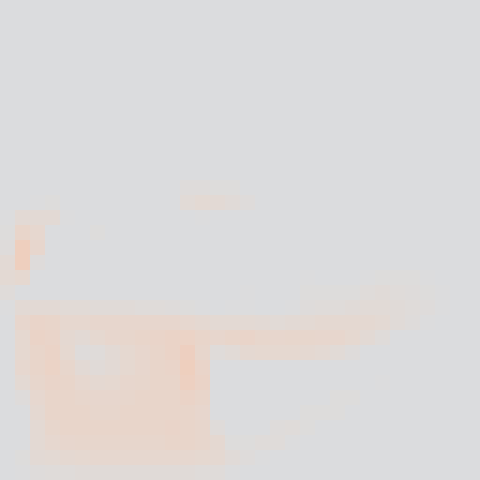} & 
\includegraphics[width = 1.4cm]{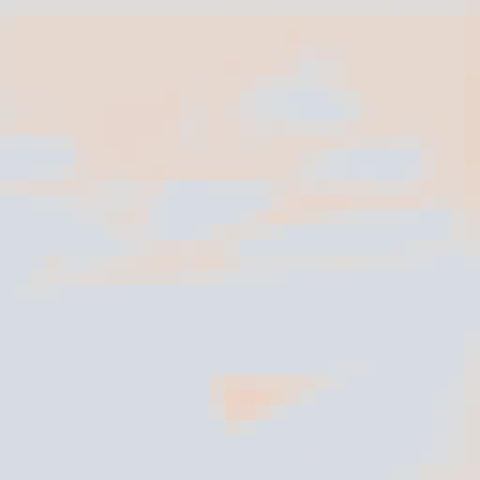} & 
\includegraphics[width = 1.4cm]{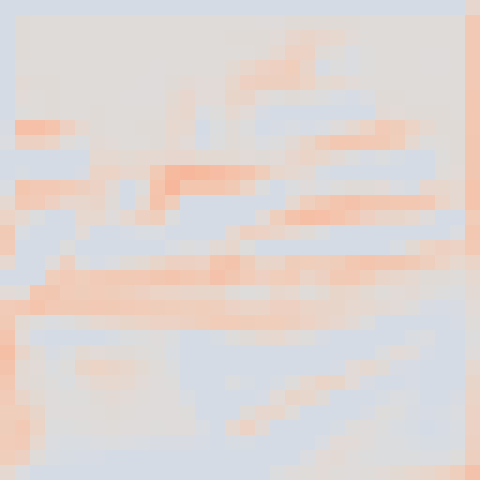} & 
\includegraphics[width = 1.4cm]{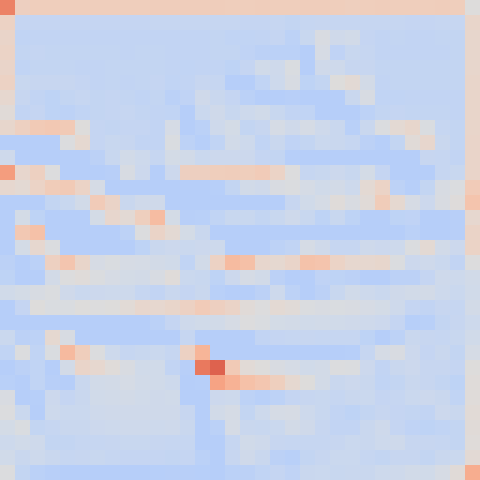} & 
\includegraphics[width = 1.4cm]{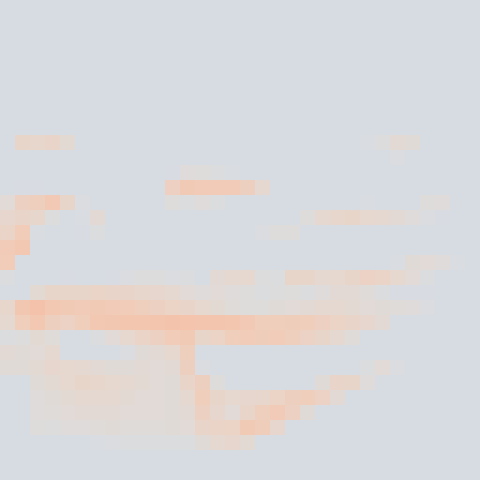} & 
\includegraphics[width = 1.4cm]{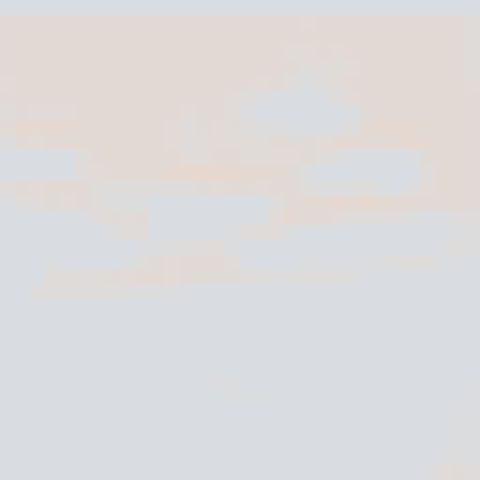} & 
\includegraphics[width = 1.4cm]{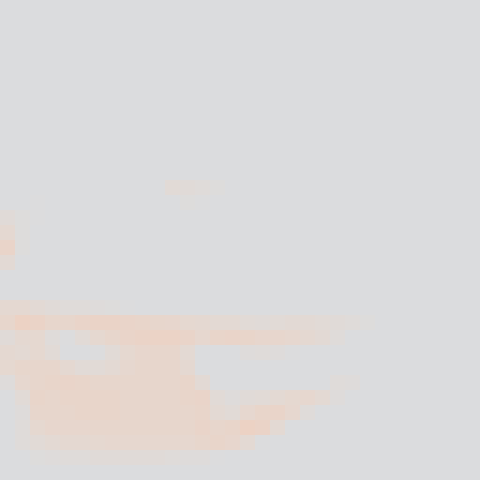} & 
\includegraphics[width = 1.4cm]{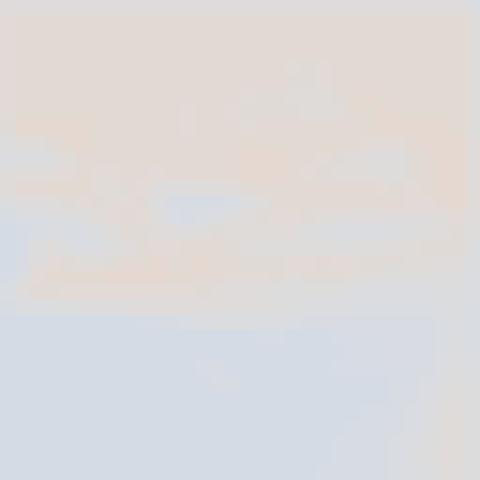}\\
16 & 17 & 18 & 19 & 20 & 21 & 22 & 23  \\

\includegraphics[width = 1.4cm]{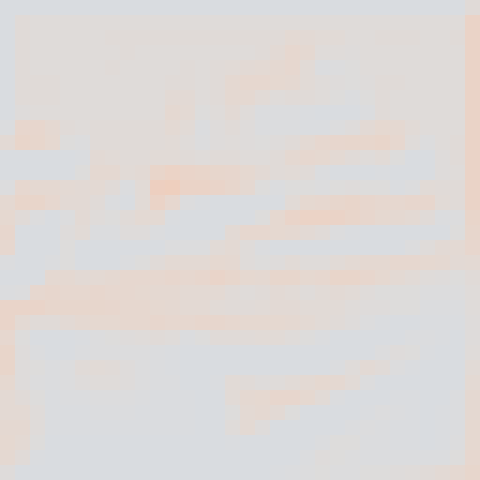} & 
\includegraphics[width = 1.4cm]{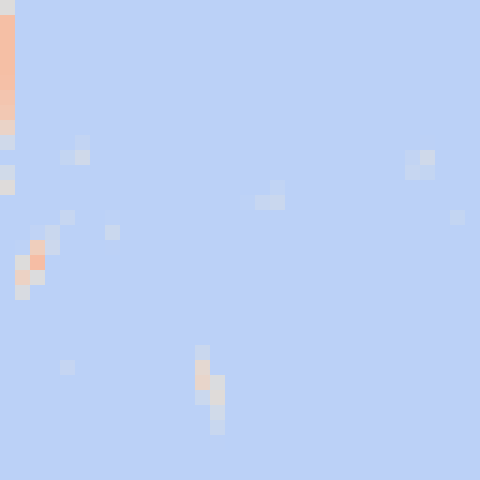} & 
\includegraphics[width = 1.4cm]{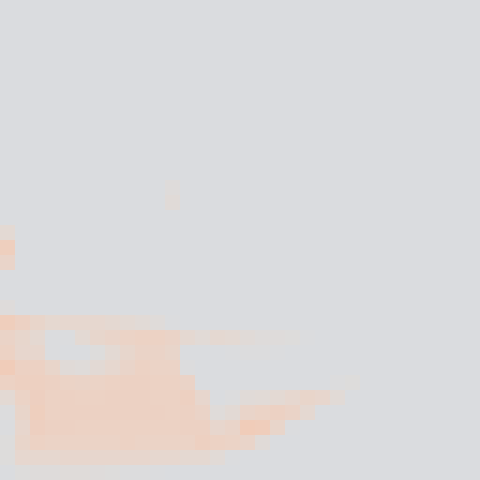} & 
\includegraphics[width = 1.4cm]{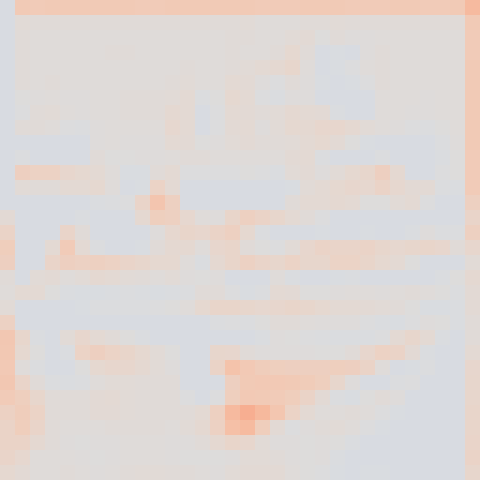} & 
\includegraphics[width = 1.4cm]{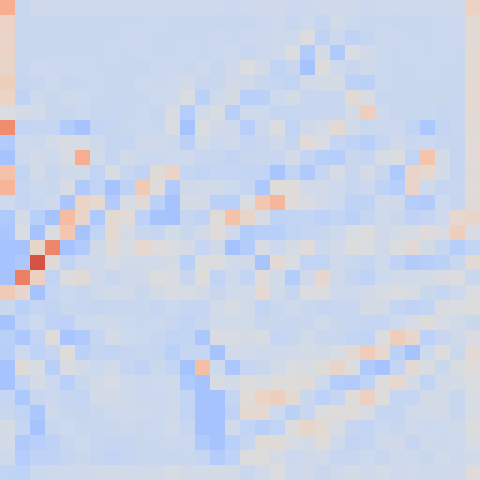} & 
\includegraphics[width = 1.4cm]{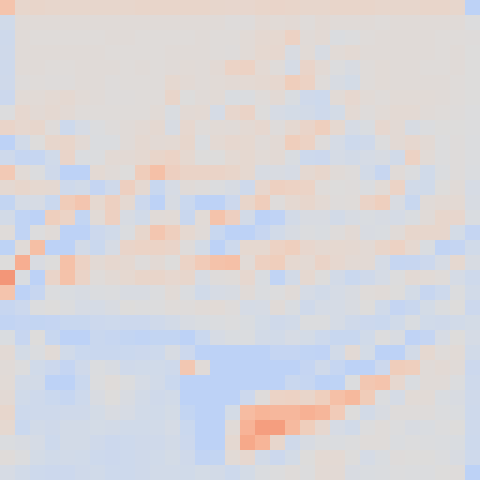} & 
\includegraphics[width = 1.4cm]{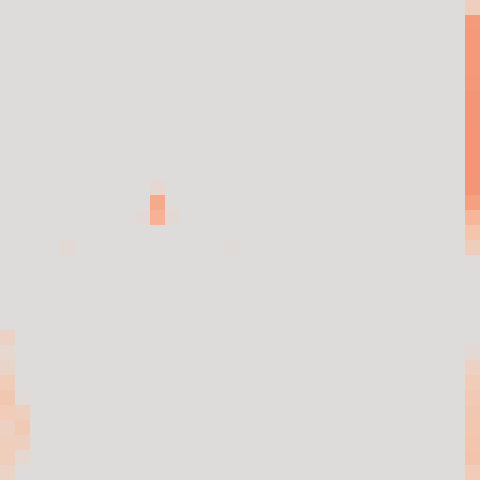} & 
\includegraphics[width = 1.4cm]{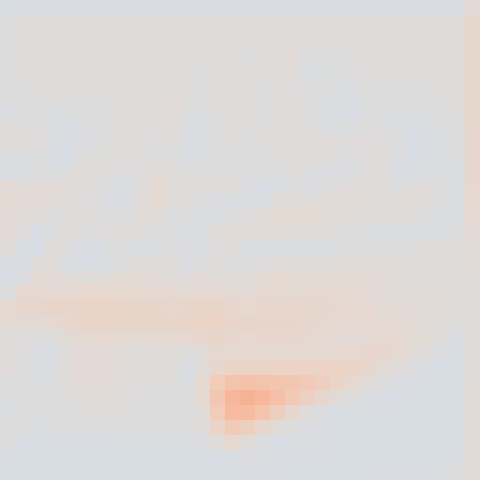}\\
24 & 25 & 26 & 27 & 28 & 29 & 30 & 31 \\

\includegraphics[width = 1.4cm]{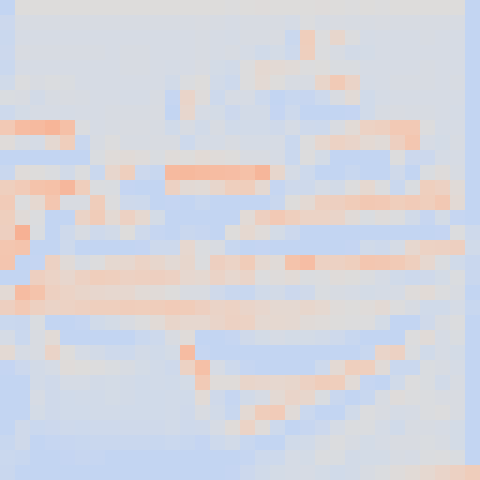} & 
\includegraphics[width = 1.4cm]{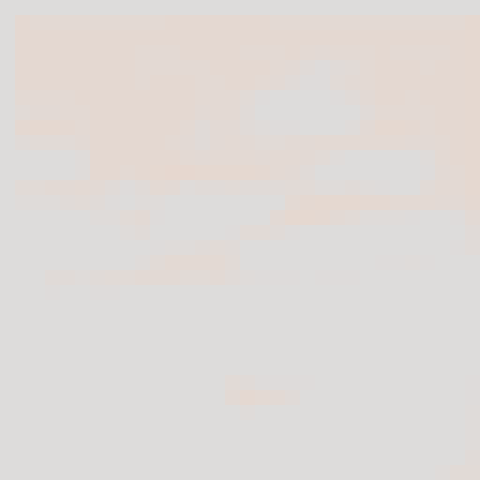} & 
\includegraphics[width = 1.4cm]{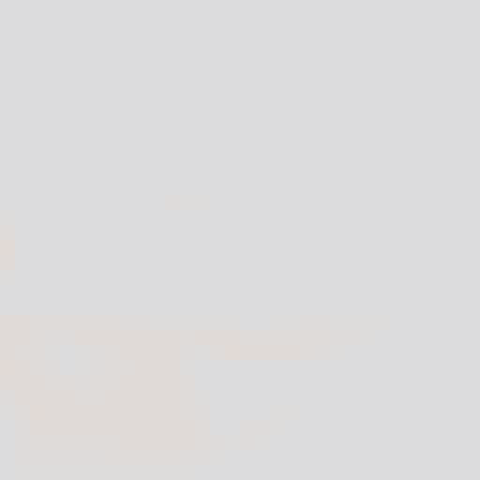} & 
\includegraphics[width = 1.4cm]{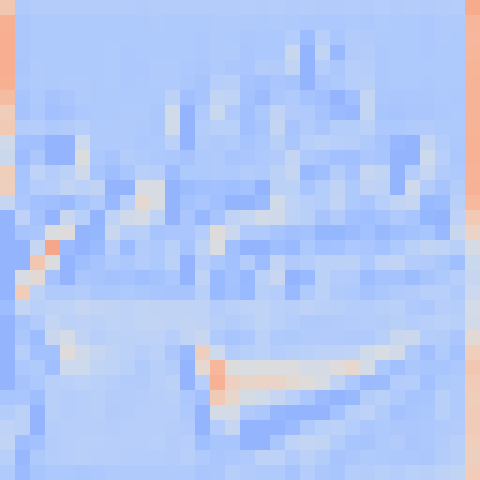} & 
\includegraphics[width = 1.4cm]{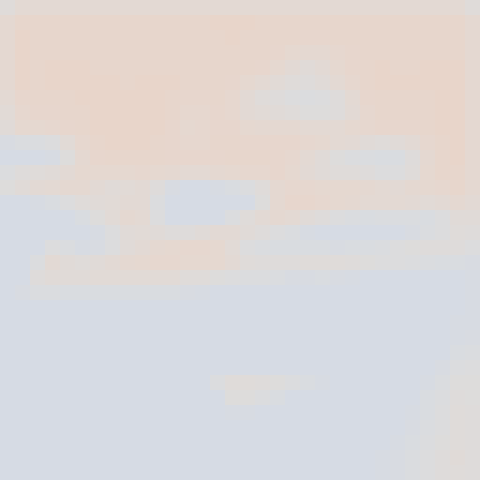} & 
\includegraphics[width = 1.4cm]{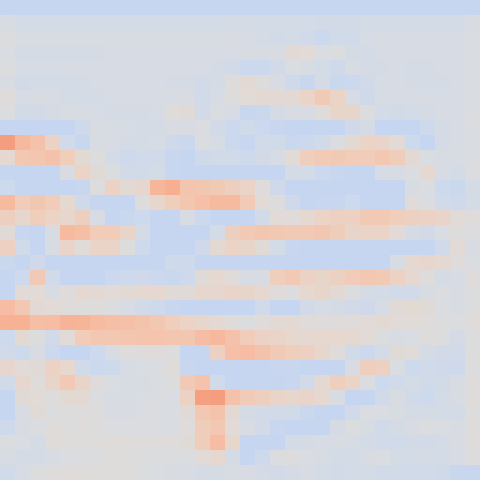} & 
\includegraphics[width = 1.4cm]{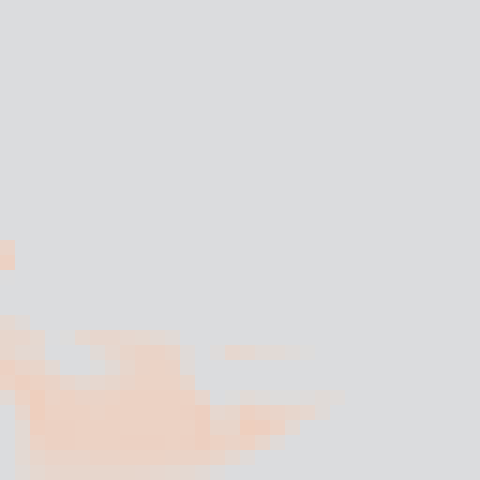} & 
\includegraphics[width = 1.4cm]{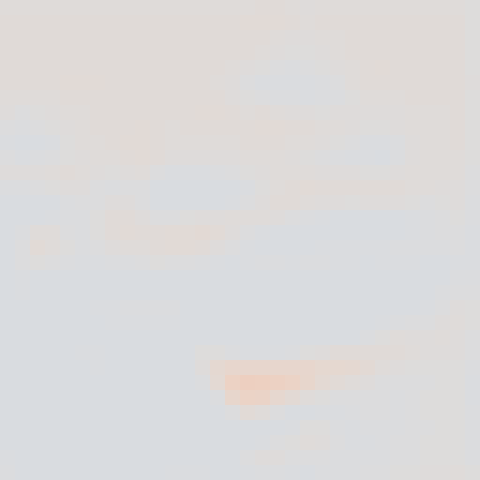}\\
32& 33 & 34 & 35 & 36 & 37 & 38 & 39  \\

\includegraphics[width = 1.4cm]{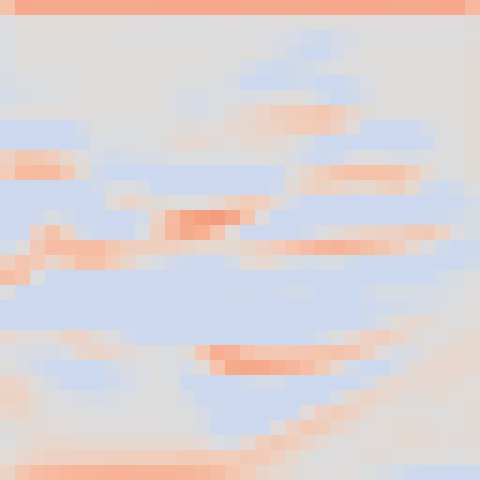} & 
\includegraphics[width = 1.4cm]{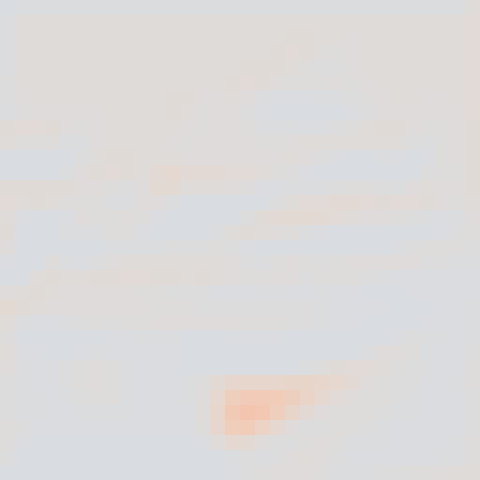} & 
\includegraphics[width = 1.4cm]{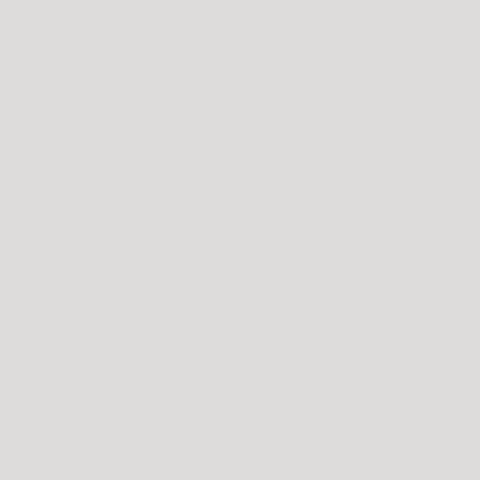} & 
\includegraphics[width = 1.4cm]{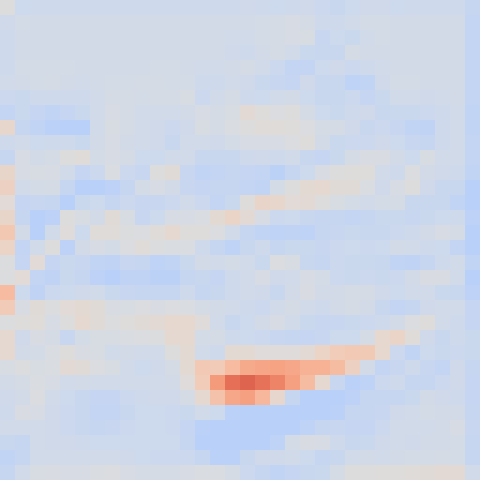} & 
\includegraphics[width = 1.4cm]{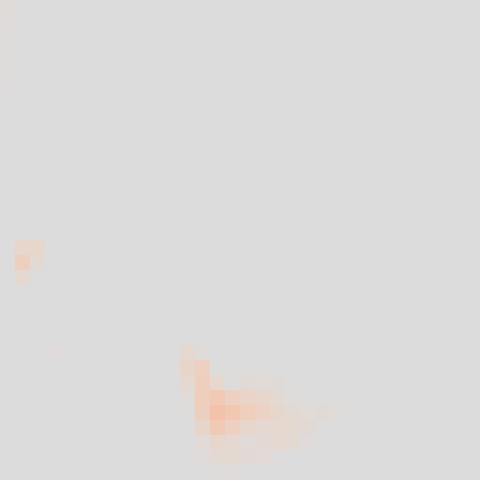} & 
\includegraphics[width = 1.4cm]{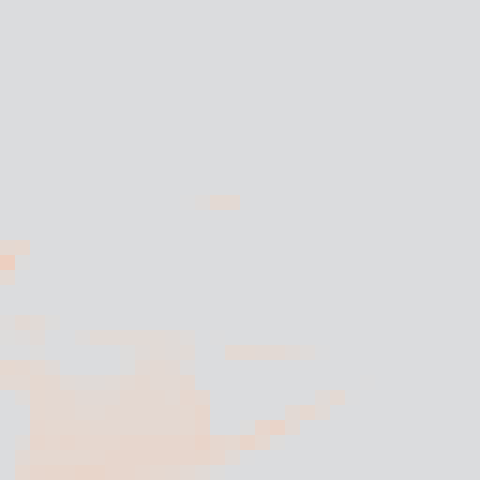} & 
\includegraphics[width = 1.4cm]{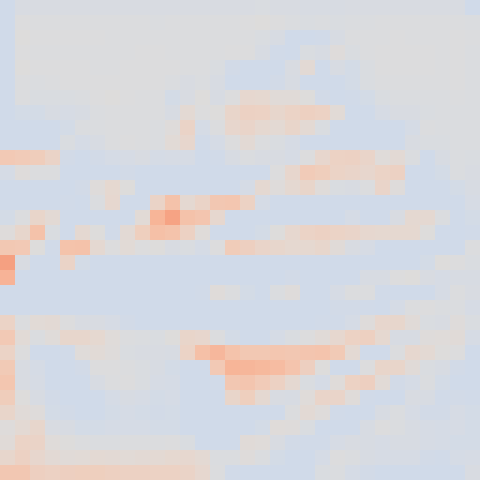} & 
\includegraphics[width = 1.4cm]{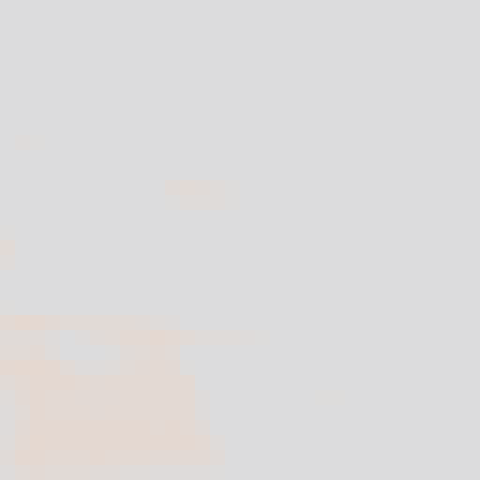}\\
40 & 41 & 42 & 43 & 44 & 45 & 46 & 47\\

\includegraphics[width = 1.4cm]{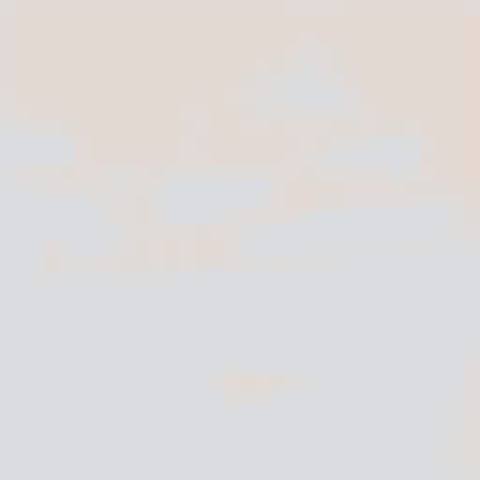} & 
\includegraphics[width = 1.4cm]{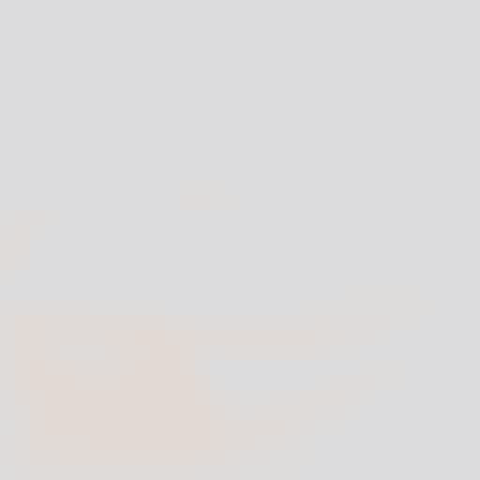} & 
\includegraphics[width = 1.4cm]{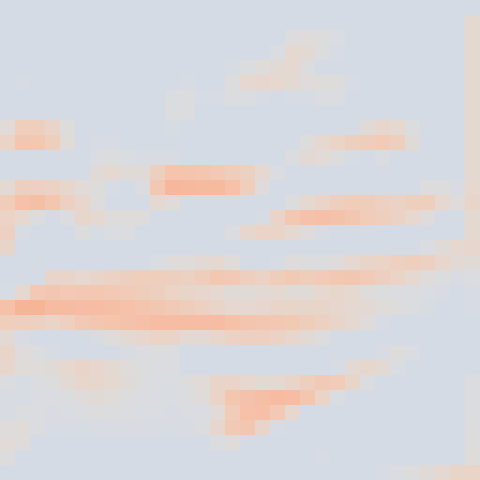} & 
\includegraphics[width = 1.4cm]{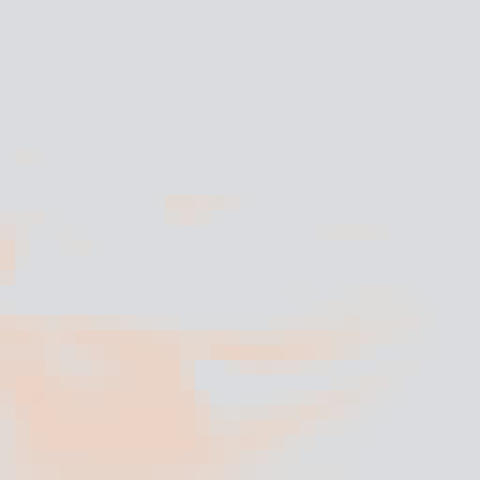} & 
\includegraphics[width = 1.4cm]{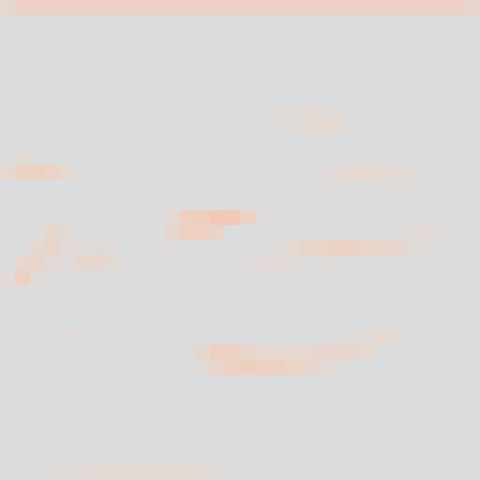} & 
\includegraphics[width = 1.4cm]{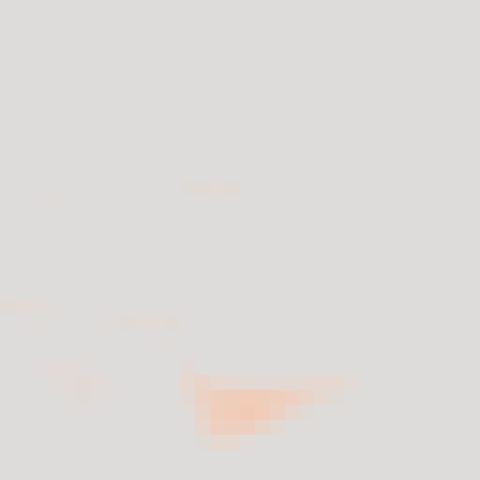} & 
\includegraphics[width = 1.4cm]{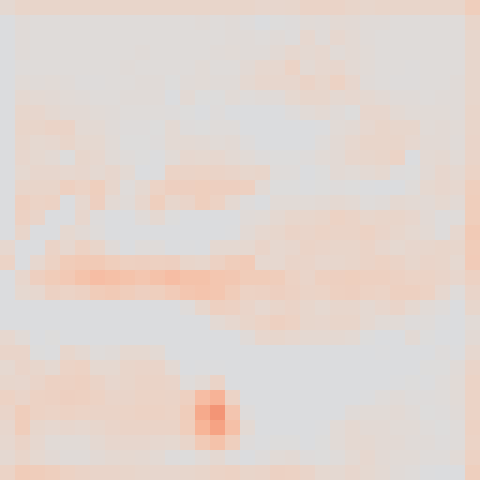} & 
\includegraphics[width = 1.4cm]{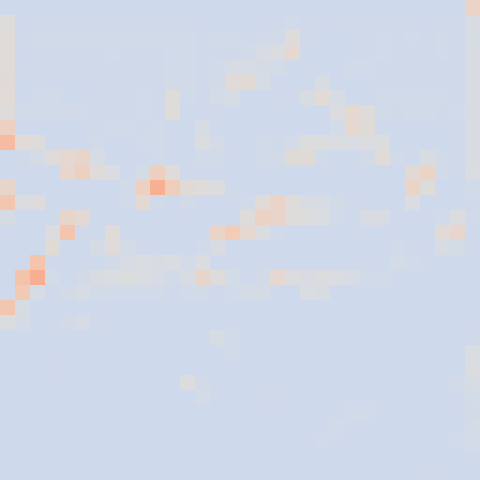}\\
48 & 49 & 50 & 51 & 52 & 53 & 54 & 55 \\

\includegraphics[width = 1.4cm]{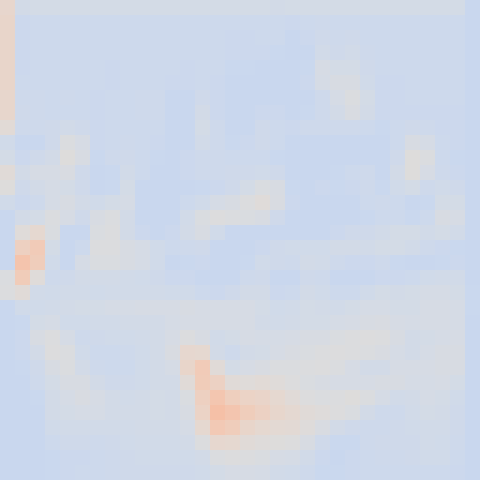} & 
\includegraphics[width = 1.4cm]{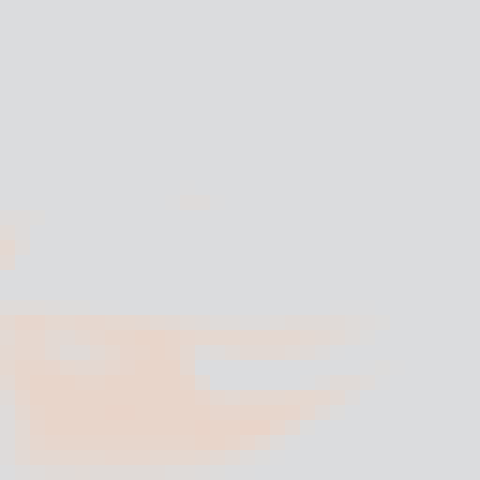} & 
\includegraphics[width = 1.4cm]{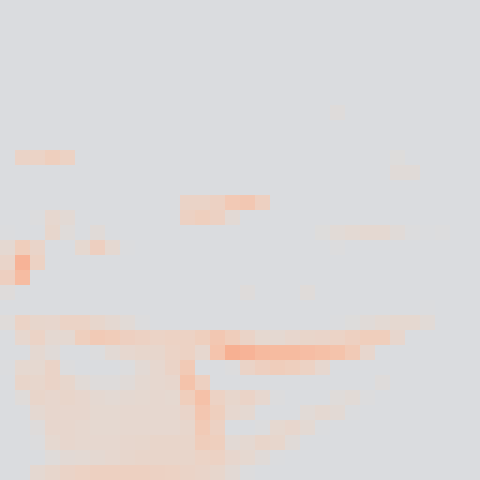} & 
\includegraphics[width = 1.4cm]{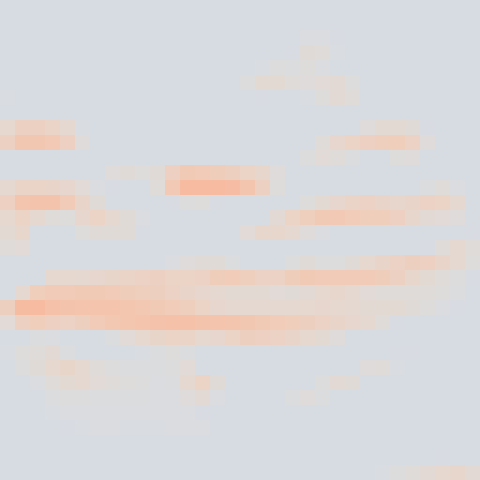} & 
\includegraphics[width = 1.4cm]{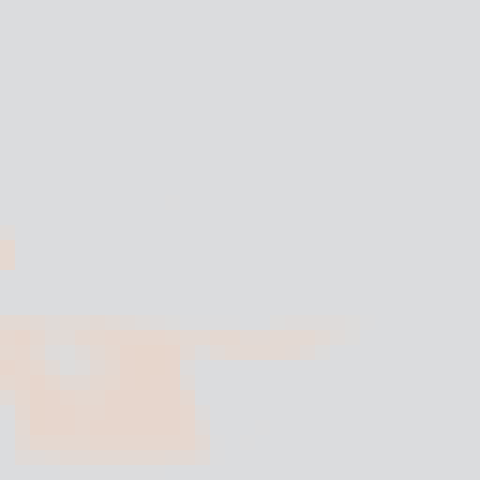} & 
\includegraphics[width = 1.4cm]{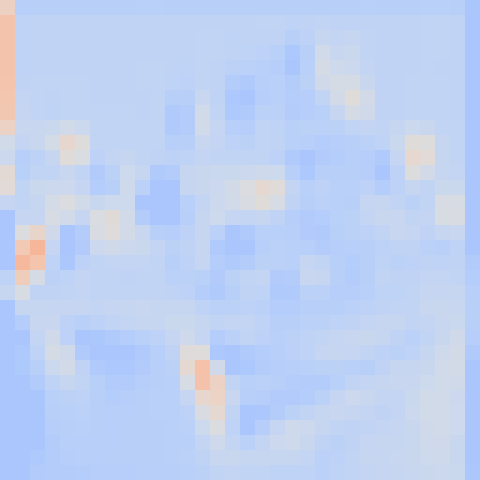} & 
\includegraphics[width = 1.4cm]{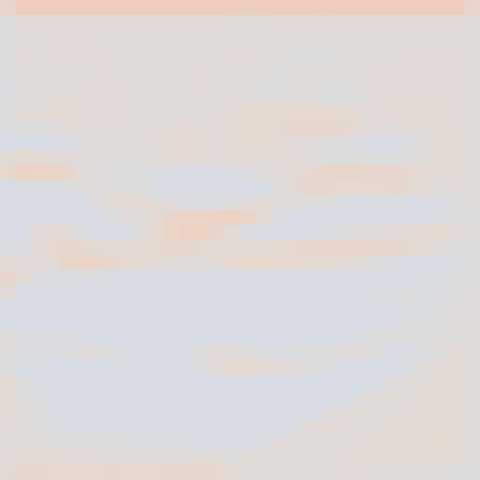} & 
\includegraphics[width = 1.4cm]{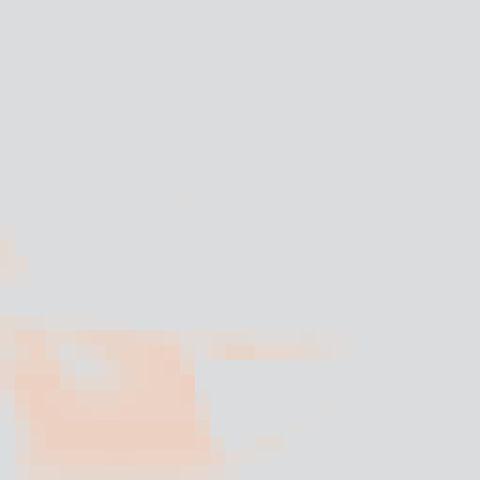}\\
56 & 57 & 58 & 59 & 60 & 61 & 62 & 63  \\

\end{tabular}
\caption{The figure presents the complete feature maps of the first convolutional layer learnt by the VGG network (presented also in Figure 1 of the main paper). The colormap indicates higher pixel values with red and lower with blue. The top filters learnt by the importance switch method are 28,15,4,29,54,9,13,3,43,6. Notice that these feature maps include the features with high activation values.}
\label{fig:allconvlayer}
\end{figure}
\end{center}

\clearpage
\bibliographystyle{plain}
\bibliography{Main}